\documentclass{article}

\usepackage{PRIMEarxiv}

\usepackage[utf8]{inputenc} 
\usepackage[T1]{fontenc}    
\usepackage{hyperref}       
\usepackage{url}            
\usepackage{booktabs}       
\usepackage{amsfonts}       
\usepackage{nicefrac}       
\usepackage{microtype}      
\usepackage{lipsum}
\usepackage{graphicx}
\usepackage{multirow}
\usepackage{xcolor}
\usepackage{pifont}
\usepackage{graphicx}
\usepackage{caption}
\usepackage{subcaption}
\usepackage{amsmath}
\newtheorem{proposition}{Proposition}[subsection]
\usepackage{hyperref}
\usepackage{natbib}
\usepackage{multirow}
\usepackage{booktabs}
\usepackage{algorithm}
\usepackage{algorithmic}
\usepackage{pdfpages}
\newcommand{\cmark}{\textcolor{green!60!black}{\ding{51}}}
\newcommand{\xmark}{\textcolor{red}{\ding{55}}}

\title{\textbf{CORP}:
\underline{C}losed-Form
\underline{O}ne-shot
\underline{R}epresentation-\underline{P}reserving
Structured Pruning}

\author{
  Boxiang Zhang, Baijian Yang\\
  Purdue University \\
  West Lafayette, IN 47907, USA\\
  \texttt{\{zhan4653, byang\}@purdue.edu} \\
}

\begin{document}
\maketitle
\footnotetext{This is a preprint. This paper has been accepted at NeurIPS 2026.\\
\hspace*{1.8em}Code is available at \url{https://github.com/BOX-LEO/CORP}.}

\begin{abstract}

Transformers~\citep{vaswani2023attentionneed} achieve strong accuracy but incur high compute and memory cost.
Structured pruning reduces inference cost, but most methods rely on retraining or multi-stage optimization, which limits post-training deployment.
We propose \textbf{CORP}, a closed-form one-shot structured pruning method that removes MLP dimensions and attention substructures using only unlabeled calibration data without gradients or fine-tuning.
CORP formulates structured pruning as a representation recovery problem.
It models removed components as affine functions of retained components and derives closed-form ridge regression solutions that fold compensation into model weights.
This minimizes a layer-local affine/logit reconstruction objective under the calibration distribution.
Experiments on ImageNet~\citep{russakovsky2015imagenetlargescalevisual} with DeiT~\citep{touvron2021trainingdataefficientimagetransformers} reveal strong redundancy in both MLP and attention representations. 
With CORP, models retain high accuracy under aggressive sparsity. 
On DeiT-Huge, CORP achieves 83.24\% Top-1 accuracy after pruning 50\% of both MLP and attention structures.

\end{abstract}

\section{Introduction}

Transformer-based models achieve strong performance across multimodal tasks but incur high computational and memory cost.
These costs limit deployment on resource-constrained hardware.
Structured pruning offers practical speedups by removing entire dimensions or substructures.
However, most structured pruning methods rely on retraining or iterative optimization.
In deployment settings, labels, gradients, and training access may not always be available.
This motivates \emph{post-training one-shot pruning}, where compression must complete in a single calibration pass without iterative optimization.
Under this constraint, naive structured pruning fails, as removing channels or attention dimensions introduces systematic error that causes rapid accuracy collapse even at moderate sparsity.

We argue that this failure cannot be explained solely by ranking quality and is largely driven by uncorrected representation error.
Structured pruning removes additive components of internal representations.
Without explicit recovery, these errors accumulate across layers.
These results suggest that compensation may be more important than increasingly complex importance metrics.

We propose \textbf{CORP}, a closed-form one-shot structured pruning framework for Transformers. CORP casts pruning as a representation recovery problem, removing MLP hidden dimensions and attention (Q/K) channels. It models removed components as affine functions of retained ones and derives closed-form ridge solutions from unlabeled calibration data. The compensation folds into weights producing standard linear layers with reduced dimensions without inference overhead.

\begin{figure*}[t]
    \centering
    \includegraphics[width=\textwidth]{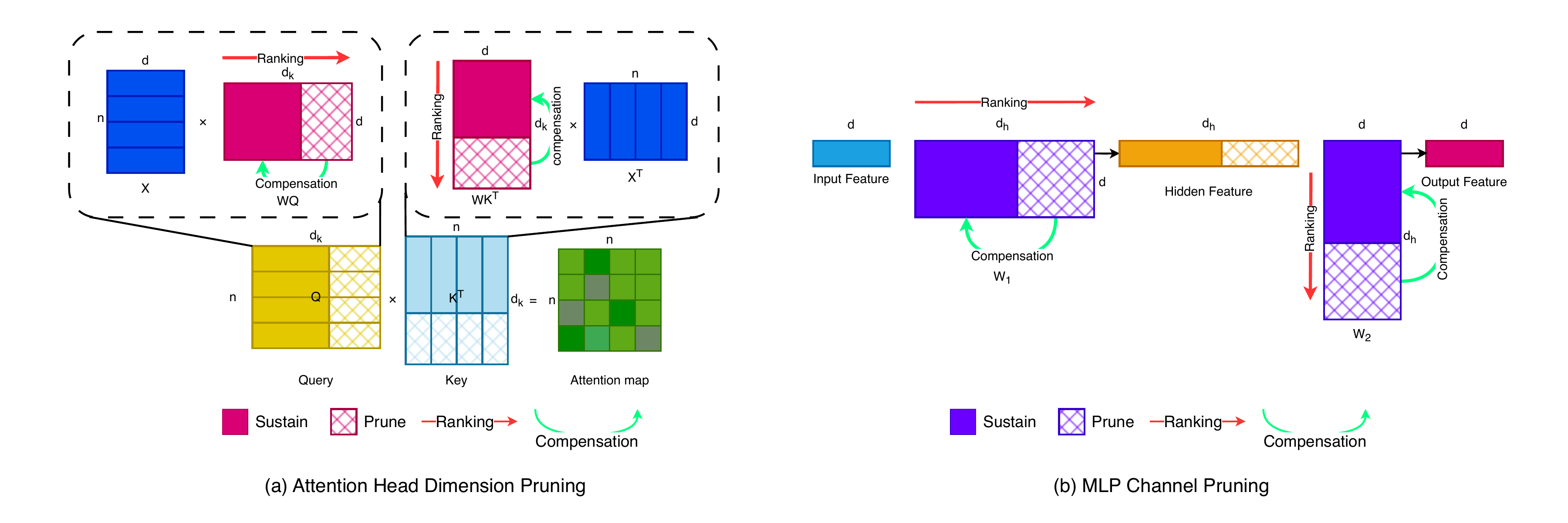}
    \caption{Illustration of CORP. Pruning removes structured components, while CORP reconstructs the missing contribution using closed-form compensation folded into the remaining weights.
    \textbf{(a)} Attention head dimension pruning removes channel dimensions in the query and key projections.
    \textbf{(b)} MLP structured pruning removes entire hidden dimensions between the two linear layers.}
    \label{fig:structured_pruning}
\end{figure*}

CORP retains strong accuracy at scale. For DeiT-Huge, it achieves 83.24\% Top-1 at 50\% sparsity with $>$40\% FLOPs/parameter reduction and 1.6$\times$ speedup on ImageNet classification. On OPT-1.3B (30\%), perplexity rises modestly (14.60$\rightarrow$18.53/19.36) with up to 17.03\% FLOPs reduction on WikiText-2 \citep{wiki_merity2016pointersentinelmixturemodels}. On DINOv2, 50\% backbone pruning preserves downstream performance, improving RMSE (0.330$\rightarrow$0.317). These results show strong one-shot structured pruning on vision models and provide preliminary evidence beyond image classification.

Our contributions are:
\begin{itemize}
\item We reframe one-shot structured pruning as a representation recovery problem, shifting focus from ranking to reconstruction, while retaining simple ranking criteria.
\item We propose CORP, a closed-form compensation framework for one-shot structured pruning without labels, gradients, or fine-tuning.
\item We demonstrate strong accuracy preservation and practical efficiency gains across scales.
\end{itemize}

\section{Related Work}
\label{sec:related}

\subsection{Post-Training Pruning}

Traditional model compression —  quantization-aware training
\citep{wang2023bitnetscaling1bittransformers,dettmers2023qlora},
pruning with fine-tuning
\citep{sanh2020movement,zhang2023dyna},
and knowledge distillation
\citep{zhao_no_2024,gou_knowledge_nodate,jin_multi-level_2023} — requires labels, gradients, or repeated optimization.
Post-training compression instead uses only a pretrained model and a small calibration set; methods differ in pruning scope, selection signal, and recovery strategy.

\paragraph{Unstructured}
Early post-training pruning focuses on unstructured pruning, which removes individual \textbf{weights}.
Optimal Brain Damage (OBD)~\citep{lecun_optimal_1989} and Optimal Brain Surgeon (OBS)~\citep{hassibi_second_1992} estimate loss sensitivity via the Hessian and compensate the model during pruning.
Post-training quantization also leverages second-order information. GPTQ~\citep{frantar_gptq_2023} uses a Hessian approximation to minimize quantization error. Optimal Brain Compression (OBC)~\citep{frantar_optimal_2023} applies layer-wise Gauss--Newton blocks to improve scalability and unify pruning and quantization under a second-order framework.

The computation of the Hessian and its inverse is the main bottleneck for scalability. WoodFisher~\citep{singh_woodfisher_2020} approximates the Hessian using empirical Fisher information with Woodbury updates. SparseGPT~\citep{frantar-sparsegpt} scales unstructured pruning to LLMs via layer-wise pruning with implicit curvature modeling.
Despite improved accuracy recovery, second-order methods require estimating or inverting large curvature matrices. These costs grow rapidly with model width and depth. Under one-shot constraints, such computation and memory overhead limit scalability to large Transformers.
Wanda~\citep{sun2024simpleeffectivepruningapproach} scales by eliminating the Hessian and ranking importance via the magnitude–activation products. While unstructured methods can achieve high sparsity with strong accuracy retention, the resulting sparse patterns yield limited inference speedups on standard hardware.

\begin{table*}[t]
\centering
\small
\resizebox{\textwidth}{!}{
\begin{tabular}{lcccc}
\toprule
Method & Pruning Scope & Selection Signal & Recovery & Finetune Free \\
\midrule
SparseGPT & Weight/N:M weight & Hessian approx. & Second-order weight update & \cmark \\
OBC & Weight/quant. group & Gauss--Newton / OBS & Second-order weight update & \cmark \\
Wanda & Weight & Act. $\times$ Mag. & None & \cmark \\
\midrule
OPTIN & MLP channel/attn head/token & Trajectory score & None & \cmark \\
Olica & MLP channel / attn head dim. & Act.\ $\times$ Mag.\ (FFN) / PCA (attn) & Low-rank ridge residual branch (FFN only) & \cmark \\
KCM & MLP channel & Output contribution & Scaling compensation & \cmark \\
VBP & MLP channel & Activation variance & Bias compensation & \xmark \\
DC-ViT & Attn module / MLP block & Recover-ability score & Iterative Feature-mimic fine-tuning & \xmark \\
GRAIL & MLP channel / attn head & Mag./Wanda/Gram selection & Post-hoc Gram-ridge reconstruction & \cmark \\
SNOWS & Mask-defined structure & Representation objective & Iterative Hessian-free recovery & \cmark \\
\midrule
CORP & MLP channel / QK dim. & Act./weight/logit energy & Closed-form affine/logit compensation & \cmark \\
\bottomrule
\end{tabular}}
\caption{Comparison of representative post-training pruning methods.
We separate pruning scope, selection signal, recovery strategy, and post-pruning
fine-tuning requirement. CORP combines MLP-channel and QK-dimension pruning with
closed-form compensation.}
\label{tab:method_compare}
\end{table*}

\paragraph{Structured Pruning}
Structured pruning removes entire heads, channels, or tokens~\citep{wang_zero-tprune_2024,peng_votemix_2024}.
However, many structured pruning methods rely on iterative mask refinement, staged optimization, or retraining, which conflict with strict one-shot post-training constraints.
Several works extend second-order objectives to structured pruning. \citet{kwon_fast_nodate} introduce Fisher-based mask variables for heads and channels. LPViT~\citep{xu_lpvit_2024} applies block-structured pruning guided by a second-order criterion.
While effective, these approaches still depend on curvature estimation and block-wise matrix inversion. 
To avoid explicit Hessian computation, OPT-IN~\citep{khaki_need_2024} use trajectory-based importance, and OFB~\citep{ye_once_nodate} learns sparsity patterns via bi-mask optimization.
LLM-oriented structured pruning differs from CORP in scope or recovery budget: FLAP~\citep{an2024flap} combines fluctuation-based global allocation with bias compensation, SliceGPT~\citep{ashkboos2024slicegpt} removes embedding dimensions via computational invariance, and LLM-Pruner~\citep{ma2023llmpruner} relies on gradient-based importance and LoRA recovery, while LPViT targets block-structured sparsity with a hardware-aware objective.

Recent work highlights the role of activation statistics. SmoothQuant~\citep{xiao_smoothquant_2024} rescales activations to suppress outliers, and AWQ~\citep{lin_awq_2024} selects important channels by activation magnitude. Olica~\citep{he2025olica} prunes FFN neurons with a Wanda-style score and reduces each head's dimension by PCA of the $W_QW_K^\top$ and $W_VW_O^\top$ products; its FFN error is recovered by a closed-form ridge fit kept as a low-rank branch at inference, with no attention compensation. TransKV~\citep{xu2026transkv} instead derives the truncation basis from data-driven keys and values. These results motivate activation-guided criteria under one-shot constraints. We call this strict single-pass, fine-tuning-free, post-training setting \emph{one-shot pruning}.

DC-ViT~\citep{zhang2024dense} performs few-shot ViT compression by removing entire attention modules and pruning portions of the remaining MLP, with a small MLP retained to absorb the resulting representation gap. Recovery relies on progressive feature-mimicking finetuning (MSE against the original model) using a tiny labeled-free training set. VBP~\citep{berisha2025variance} performs structured MLP pruning using activation statistics without gradients or Hessian, but relies on ten epochs of distillation fine-tuning to recover accuracy. KCM~\citep{nova_gradient-free_2023} ranks MLP channels by layer-wise output statistics and mutual representability without gradients or fine-tuning, then rescales the kept channels to reconstruct layer-wise outputs.
SNOWS~\citep{lucas_preserving_2024} preserves deeper representations using Hessian-free second-order optimization. 
GRAIL~\citep{tang2026grailposthoccompensationlinear} performs post-hoc linear reconstruction after structured compression. 
These methods also recover representations after pruning, but differ from CORP in pruning scope, recovery objective, optimization strategy, and finetuning dependency (Table~\ref{tab:method_compare}).

\subsection{Positioning}
CORP differs from prior representation-recovery methods in three ways: it jointly prunes MLP channels and Q/K attention dimensions, derives a closed-form logit-space compensator for attention, and folds all compensation directly into reduced projections without iterative optimization or extra inference modules.
SNOWS~\citep{lucas_preserving_2024} also performs one-shot post-training pruning without retraining, but relies on Hessian-free second-order optimization to preserve deeper nonlinear representations. In contrast, CORP uses local closed-form ridge compensation and avoids gradients, Hessian-vector products, and iterative updates.
Recent concurrent work, GRAIL~\citep{tang2026grailposthoccompensationlinear}, is closer in spirit. For MLP blocks, both GRAIL and CORP use linear reconstruction after structured pruning. However, GRAIL compensates only the second linear layer to recover module outputs, while CORP performs layer-wise representation recovery and additionally introduces Q/K logit-space compensation for attention-dimension pruning by approximating $Q_P K_P^\top$ with $Q_S M K_S^\top$ and folding the correction into the query and key projections.
Olica also fits a closed-form ridge solution, but regresses the pruned FFN's output residual on the block input and keeps it as an extra low-rank branch; CORP predicts the pruned activations from the retained ones and folds the correction into the existing MLP and Q/K weights, adding no inference-time parameters.
Thus, CORP unifies MLP-channel and Q/K-dimension pruning under a single closed-form compensation framework.
Like L-OBS~\citep{dong2017lobs}, AdaQuant~\citep{hubara2021adaquant}, OBC, GPTQ, SparseGPT, and KCM, CORP preserves each layer's input--output mapping from a small unlabeled calibration set; it differs in using closed-form compensation for MLP channels and Q/K dimensions.

\section{Method}

\subsection{Empirical Motivation and Hypothesis}

We empirically find that MLP blocks in DeiT-Base \citep{touvron2021trainingdataefficientimagetransformers} are highly over-parameterized. 
Detailed statistics are reported in Table~\ref{tab:mlp_redundancy} (Appendix~\ref{app:mlp_redundancy}). 
However, naive structured pruning causes severe accuracy degradation. 
We hypothesize that structured pruning introduces a \emph{systematic} error: the structured, mask-determined contribution that pruning deletes from a module output, namely $W_Px_P$ for an MLP block and $Q_PK_P^\top$ for an attention head (Sec.~\ref{sec:notation}).
This term is not the total prediction error but the exact residual created by the mask; objectives~\eqref{eq:mlp_obj} and~\eqref{eq:attn_obj} minimize its expected squared norm, i.e., they recover each module's output rather than optimize the end-to-end task loss (see the decompositions in Eqs.~\eqref{eq:mlp-shift-main} and~\eqref{eq:attn-shift-main}).
Correcting this error preserves the original representation.
Based on this hypothesis, we cast pruning as a representation approximation problem. 
We remove redundant channels and compensate the induced error with a lightweight affine transformation. 
This yields a closed-form correction that preserves intermediate representations.
We treat structured pruning as representation recovery on top of importance-based selection.

\subsection{Problem Setup and Notation}
\label{sec:notation}

We consider one-shot structured pruning of a pretrained Transformer
using a small calibration set $\mathcal{D}=\{x^{(i)}\}_{i=1}^N$.
All expectations are estimated empirically over $\mathcal{D}$.

\paragraph{MLP} consists of two linear layers. The first projects activations to a higher-dimensional space, and the second maps them back to the original dimension. Our goal is to reduce the hidden dimension.
We model the second MLP layer as an affine map defined by $W \in \mathbb{R}^{d \times o}$ and $b \in \mathbb{R}^{d}$ where $d$ is the feature dimension and $o$ is the hidden dimension:
\begin{equation}
y = Wx + b, \quad W=[W_S\ \ W_P],\quad x=\begin{bmatrix}x_S\\ x_P\end{bmatrix}.
\end{equation}
Structured pruning removes input channels by partitioning indices into kept set $S$ and pruned set $P$.
After pruning, only $x_S$ is available.
We seek compensated parameters $(\widehat W_S,\widehat b)$ that minimize
\begin{equation}
\min_{\widehat W_S,\widehat b}
\ \mathbb{E}_{x\sim\mathcal{D}}
\bigl\|Wx+b-(\widehat W_S x_S+\widehat b)\bigr\|_2^2.
\label{eq:mlp_obj}
\end{equation}
\paragraph{Attention}
For self-attention, let $X\in\mathbb{R}^{n\times d}$ denote the token matrix.
Query and key projections are
\begin{equation}
Q = XW_Q + \mathbf{1}b_Q^\top,\quad
K = XW_K + \mathbf{1}b_K^\top,
\end{equation}
with $W_Q,W_K\in\mathbb{R}^{d\times d_h}$, where $d_h$ is the head dimension.
We prune by splitting head dimensions.
\begin{equation}
Q=[Q_S\ Q_P], K=[K_S\ K_P] \quad \Rightarrow\quad L = QK^\top = Q_S K_S^\top + Q_P K_P^\top.
\end{equation}

We seek compensated parameters $(\widehat Q_S,\widehat K_S)$ that minimize
\begin{equation}
\min_{\widehat Q_S,\widehat K_s}
\ \mathbb{E}_{x\sim\mathcal{D}}
\bigl\|QK^\top-(\widehat Q_S\widehat K_S^\top)\bigr\|_F^2.
\label{eq:attn_obj}
\end{equation}

We restrict attention pruning to the \textbf{Query} and \textbf{Key} projections and do not prune the Value projection.
In multi-head attention, 
pruning $V$ is more coupled. 
Removing $V$ dimensions changes the per-head features before concatenation. Recovering it requires joint modeling of the attention weights, per-head value features, and the post-concatenation output projection. That is beyond the scope of this study.


\subsection{Ranking} \label{sec:ranking}
We rank MLP channels using simple, data-driven metrics based on activation energy and weight magnitude. For each channel $i$, the activation energy is defined as $E_i = \mathbb{E}_{x \sim \mathcal{D}}\left[x_i^2\right]$, computed on the calibration set $\mathcal{D}$. We also consider the corresponding weight magnitude $|W{:,i}|_2$, capturing its impact on the layer output. Inspired by Wanda, we also evaluate a combined metric $E_i |W{:,i}|_2$. Channels are ranked by these scores and those with the lowest importance are pruned. 
We rank head dimensions by  expected logit energy.
For column $j$ of $Q$ and $K$, the energy is $s_j = \mathbb{E}\!\left[\|q_j\|_2^2\,\|k_j\|_2^2\right].$
Dimensions with the smallest $s_j$ are pruned.

\subsection{Closed-form Compensation}
\paragraph{MLP}
\label{sec:mlp_affine}

We predict pruned activations from retained ones and fold the correction into the remaining weights, compensating the output error introduced by pruning. The original output is
\begin{equation}
y = W_S x_S + W_P x_P + b.
\end{equation}

We approximate pruned activations as an affine function of kept activations:
\begin{equation}
x_P \approx Bx_S + c,
\end{equation}

We estimate $(B,c)$ via ridge regression:
\begin{equation}
    \min_{B,c}\ \|X_P-(BX_S+c\mathbf{1}^\top)\|_F^2+\lambda\|B\|_F^2,
\label{eq:mlp_ridge}
\end{equation}

The closed-form solution is:
\begin{equation}
B = \Sigma_{PS}(\Sigma_{SS}+\lambda I)^{-1},\qquad
c = \mu_P - B\mu_S.
\label{eq:mlp_closed}
\end{equation}
\begin{equation}
\Sigma =
\begin{bmatrix}
\Sigma_{SS} & \Sigma_{SP} \\
\Sigma_{PS} & \Sigma_{PP}
\end{bmatrix},
\quad
\Sigma_{SS}=\mathbb E[(x_S-\mu_S)(x_S-\mu_S)^\top],
\quad
\Sigma_{PS}=\mathbb E[(x_P-\mu_P)(x_S-\mu_S)^\top].
\label{eq:mlp_cov}
\end{equation}
A full derivation of the affine predictor and its closed-form solution is provided in Appendix~\ref{app:closed_form_mlp}.


We next bound the layer distortion; the full derivation appears in Appendix~\ref{app:bounds-mlp}. The distortion equals:
\begin{equation}
J_{\mathcal D}^\star
\;=\;
\mathrm{tr}\!\bigl( W_P\,\Sigma_{P|S}\,W_P^\top \bigr),
\qquad
\Sigma_{P|S} := \Sigma_{PP} - \Sigma_{PS}\,\Sigma_{SS}^\dagger\,\Sigma_{SP},
\label{eq:mlp-exact-main}
\end{equation}
Comparing to the uncompensated baseline
$J_{\mathrm{uncomp}} = \mathrm{tr}(W_P\,\Sigma_{PP}\,W_P^\top) + \|W_P\mu_P\|_2^2$ yields:
\begin{equation}
J_{\mathrm{uncomp}} - J_{\mathcal D}^\star
\;=\;
\underbrace{\mathrm{tr}\!\bigl(W_P\,\Sigma_{PS}\,\Sigma_{SS}^\dagger\,\Sigma_{SP}\,W_P^\top\bigr)}_{\geq\,0}
\;+\;
\underbrace{\|W_P\,\mu_P\|_2^2}_{\geq\,0},
\label{eq:mlp-gain-main}
\end{equation}
so compensation strictly reduces the error whenever the kept activations carry any linear signal about the pruned ones in the directions of $W_P$,
or the pruned channels have non-zero mean.

In practice CORP fits $(\widehat B_\lambda,\widehat c_\lambda)$ on a
calibration distribution $\mathcal D$ from $N$ samples and deploys on a
distribution $\mathcal D_t$. The test-time error decomposes as
\begin{equation}
\mathcal E_t(\widehat B_\lambda,\widehat c_\lambda)
\;\leq\;
2\,J_{\mathcal D_t}^\star
\;+\; \mathcal{O}\!\bigl( |S|/N \bigr)
\;+\; \mathcal{O}\!\bigl( \|\Sigma^{(c)} - \Sigma^{(t)}\|_{\mathrm{op}} \bigr),
\label{eq:mlp-shift-main}
\end{equation}
The three terms isolate (i)~the irreducible affine residual on $\mathcal D_t$, (ii)~the calibration sample-complexity error, and (iii)~the calibration--deployment shift.
Equations~\eqref{eq:mlp-exact-main}--\eqref{eq:mlp-shift-main} make the
two roles of CORP explicit: ranking controls the magnitude of $W_P$, while
compensation controls the residual covariance $\Sigma_{P|S}$. The method does not require globally low-rank activations; only the conditional residual
$\Sigma_{P|S}$ in the directions of $W_P$ must be small.

\paragraph{Attention}
\label{sec:attn_comp}
To compensate for the missing logits, we approximate
$Q_P K_P^\top \approx Q_S M K_S^\top$, 
where $M\in\mathbb{R}^{d_h'\times d_h'}$ models interaction between removed and kept logit subspaces for each head. 
For sample $b$ in calibration set $\mathcal D$, we estimate $M$ for each layer and attention head by solving
\begin{equation}
\min_M
\sum_{b=1}^{N}
\|Q_{P,b}K_{P,b}^{\top}-Q_{S,b}MK_{S,b}^{\top}\|_F^2
+\lambda\|M\|_F^2 .
\label{eq:attn_ridge_obj}
\end{equation}
Vectorizing gives the closed-form ridge system
\begin{equation}
\left[
\sum_{b=1}^{N}
(K_{S,b}^{\top}K_{S,b})\otimes(Q_{S,b}^{\top}Q_{S,b})
+\lambda I
\right]\mathrm{vec}(M)
=
\sum_{b=1}^{N}
\mathrm{vec}\left(
(Q_{S,b}^{\top}Q_{P,b})(K_{P,b}^{\top}K_{S,b})
\right).
\label{eq:attn-ridge1}
\end{equation}
The derivation of the normal equations and logit-space compensation is given in Appendix~\ref{app:closed_form_attn}.
Finally, we factor $I+M=U\Sigma V^\top$ and fold compensation into the projections:
\begin{equation}
\widehat W_{Q,S}=W_{Q,S}U\Sigma^{1/2},\qquad
\widehat W_{K,S}=W_{K,S}V\Sigma^{1/2}.
\end{equation}



Let $\widehat M_\lambda$ be the ridge-compensated solution
in~\eqref{eq:attn-ridge1}, and substitute $G = \sum_b (K_{S,b}^\top K_{S,b})\otimes(Q_{S,b}^\top Q_{S,b})$, $h = \sum_b \mathrm{vec}\bigl((Q_{S,b}^\top Q_{P,b})(K_{P,b}^\top K_{S,b})\bigr)$ into the objective gives the unregularized optimum distortion
\begin{equation}
J_{\mathcal D}^\star
\;=\; \sum_{b=1}^{N}\! \| Q_{P,b} K_{P,b}^\top \|_F^2
\;-\; h^\top G^\dagger\, h,
\label{eq:attn-exact-main}
\end{equation}
where the first term is the uncompensated logit error obtained by
$M=0$ and the second is the strict compensation gain.
The gain is non-negative,
and equals zero only when the kept Q/K subspaces carry no bilinear signal about the pruned ones. 

Under calibration--test shift, the test-time logit distortion decomposes as
\begin{equation}
\mathcal E_t(\widehat M_\lambda)
\;\leq\;
2\,J_{\mathcal D_t}^\star
\;+\; \mathcal{O}\!\bigl( d_h'^2 / N \bigr)
\;+\; \mathcal{O}\!\bigl( \|G^{(c)}/N - G^{(t)}\|_{\mathrm{op}} \bigr),
\label{eq:attn-shift-main}
\end{equation}
The three terms have the same interpretation as in the MLP case, with the parameter dimension $d_h'^2$, the size of the per-head compensator $M$.
A full derivation of the bound is provided in Appendix~\ref{app:bounds-attn}.
These bounds justify the local compensation objective for deployment data close to the calibration distribution, which Appendix~\ref{app:rho2} verifies on held-out images and classes; they do not imply global optimality or robustness under arbitrary shift.
This locality is a deliberate trade-off. Given the activations cached from one pass of the unpruned model, each objective is quadratic in its own compensator, so every MLP block and attention head admits a closed-form ridge solution and all solves run independently and in parallel, without labels, gradients, or sequential re-calibration (compensation is below 1\% of total runtime, Table~\ref{tab:runtime}). Coupling modules through the intervening nonlinearities would make the objective non-quadratic and require iterative optimization, as in SNOWS~\citep{lucas_preserving_2024}. The price is that the residual $e_\ell$ left by each module is passed downstream unmodeled. The deviation of the final representation is bounded by a sum of the local residual norms weighted by the operator norms of the downstream Jacobians, as in layer-wise post-training compression~\citep{dong2017lobs,hubara2021adaquant,frantar_optimal_2023,frantar_gptq_2023,frantar-sparsegpt,nova_gradient-free_2023}; minimizing each $\|e_\ell\|$ therefore controls, but does not eliminate, accumulated error, since the bound is an upper bound and deeper layers inherit the residuals of earlier ones.

\begin{algorithm}[h]
\caption{CORP: One-shot Structured Pruning with Compensation}
\label{alg:CORP}
\begin{algorithmic}
\STATE {\bfseries Input:} Pretrained model $f_\theta$, calibration set $\mathcal{D}$,
MLP sparsity $s_{\mathrm{mlp}}$, attention sparsity $s_{\mathrm{attn}}$, ridge $\lambda$
\STATE {\bfseries Output:} Pruned and compensated model $f_{\hat\theta}$

\STATE Run $f_\theta$ on $\mathcal{D}$ and cache MLP activations and attention $Q,K$

\FOR{each MLP block, each attention head}
  \STATE Rank hidden channels  \textsc{RankMLP} (Alg.~\ref{alg:rank_mlp})
  \STATE Rank $Q/K$ dimensions \textsc{RankAttn} (Alg.~\ref{alg:rank_attn})
  \STATE  affine compensation  \textsc{CompensateMLP} (Alg.~\ref{alg:comp_mlp})
  \STATE  logit compensation  \textsc{CompensateAttn} (Alg.~\ref{alg:comp_attn})
\ENDFOR

\STATE {\bfseries return} $f_{\hat\theta}$
\end{algorithmic}
\end{algorithm}

Algorithm~\ref{alg:CORP} summarizes the complete CORP pipeline.
All statistics are collected in a single forward pass of the \emph{unpruned} model: $X_S,X_P$ in Eq.~\eqref{eq:mlp_ridge} and $Q_S,Q_P,K_S,K_P$ in Eq.~\eqref{eq:attn_ridge_obj} are taken from the original network, with subscripts only indexing kept and pruned subsets. Each MLP layer and attention head is then solved independently from this one cache, which is what allows a single calibration pass and parallel solves; there is no joint or sequential optimization across modules.

\begin{table*}[t]
\centering
\footnotesize
\setlength{\tabcolsep}{3pt}
\renewcommand{\arraystretch}{1.1}
\resizebox{\textwidth}{!}{%
\begin{tabular}{lcccccccccc}
\toprule
 & Base & \multicolumn{3}{c}{MLP} & \multicolumn{3}{c}{Attn} & \multicolumn{3}{c}{Both} \\
\cmidrule(lr){2-2}
\cmidrule(lr){3-5}
\cmidrule(lr){6-8}
\cmidrule(lr){9-11}
Model 
& Top1 $\mid$ FLOPs $\mid$ P
& Top1 & FLOPs  $\mid \downarrow$ & P
& Top1 & FLOPs  $\mid \downarrow$ & P
& Top1 & FLOPs  $\mid \downarrow$ & P \\
\midrule
Tiny  
& 72.02 $\mid$ 1.4 $\mid$ 5.7
& 55.43$\pm$0.05 & 1.1 $\mid$ 24.3\% & 3.9 
& 64.33$\pm$0.14 & 1.1 $\mid$ 21.6\% & 5.3 
& 41.69$\pm$0.24 & 0.8 $\mid$ 45.9\% & 3.5 \\

Small 
& 79.72 $\mid$ 5.0 $\mid$ 22.1
& 69.35$\pm$0.12 & 3.6 $\mid$ 28.1\% & 15.0 
& 72.52$\pm$0.06 & 4.2 $\mid$ 16.0\% & 20.3 
& 58.47$\pm$0.26 & 2.8 $\mid$ 44.1\% & 13.2 \\

Base  
& 81.74 $\mid$ 18.3 $\mid$ 86.6
& 72.20$\pm$0.16 & 12.7 $\mid$ 30.5\% & 58.2 
& 80.76$\pm$0.04 & 16.0 $\mid$ 12.5\% & 79.5 
& 70.17$\pm$0.14 & 10.4 $\mid$ 43.0\% & 51.2 \\

Large 
& 84.58 $\mid$ 63.5 $\mid$ 304.4
& 82.06$\pm$0.06 & 43.7 $\mid$ 31.2\% & 203.7 
& 83.62$\pm$0.04 & 56.2 $\mid$ 11.6\% & 279.2 
& 80.34$\pm$0.06 & 36.3 $\mid$ 42.8\% & 178.5 \\

Huge  
& 84.97 $\mid$ 173.0 $\mid$ 632.1
& 84.09$\pm$0.04 & 119.0 $\mid$ 31.2\% & 422.3 
& 84.22$\pm$0.03 & 153.0 $\mid$ 11.7\% & 579.7 
& 83.24$\pm$0.05 & 98.7 $\mid$ 42.9\% & 369.9 \\
\bottomrule
\end{tabular}}
\caption{This table reports Top-1 accuracy (Top1), FLOPs in billions (G), and parameters in millions P under 50\% structured sparsity. For pruned models, FLOPs are presented as `FLOPs $\mid \downarrow$'', where $\downarrow$ indicates the percentage reduction relative to the corresponding base model. Top-1 of pruned models is reported as mean $\pm$ sample standard deviation over five independently sampled calibration sets; FLOPs and parameter counts are identical across these runs.}
\label{tab:corp_full}
\end{table*}
\section{Experiments}
\label{sec:exp}

We evaluate pretrained DeiT~\citep{touvron2021trainingdataefficientimagetransformers,touvron2022deitiiirevengevit} and ViT~\citep{dosovitskiy2021imageworth16x16words} on ImageNet~\citep{russakovsky2015imagenetlargescalevisual} validation set.
For transferability, we prune pretrained DINOv2 \citep{oquab2024dinov2learningrobustvisual} backbones and evaluate them on NYU Depth V2 \citep{nyuv2} and ADE20k \citep{ADE20K}. 
For language modeling, we prune a pretrained OPT \citep{zhang2022optopenpretrainedtransformer} model using calibration samples from C4 \citep{c4_dodge-etal-2021-documenting} and report perplexity on WikiText2 \citep{wiki_merity2016pointersentinelmixturemodels}.
All experiments run on a single 24GB NVIDIA RTX 3090 GPU  without kernel fusion or CUDA optimizations.

\subsection{Results}
\begin{figure}[t]
\centering

\begin{minipage}{0.52\linewidth}
    \centering
    \includegraphics[width=\linewidth]{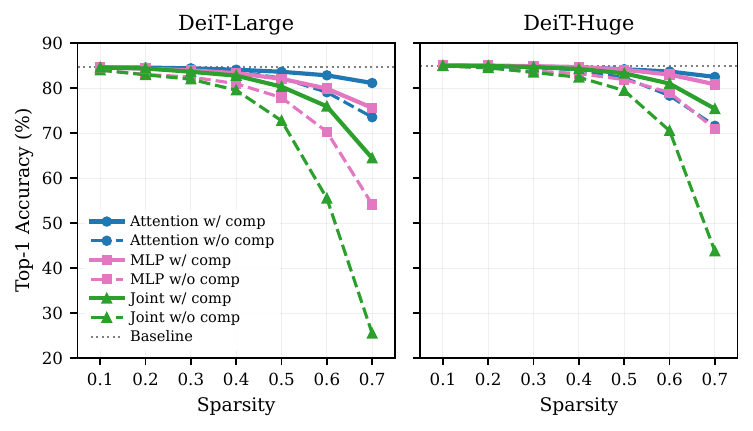}
    \caption{Top-1 ImageNet accuracy vs. sparsity. 
    }
    \label{fig:acc_sparsity}
\end{minipage}
\hfill
\begin{minipage}{0.45\linewidth}
\centering
\small
\setlength{\tabcolsep}{4pt}
\begin{tabular}{lccccc}
\toprule
\textbf{Calib} & \textbf{Tiny} & \textbf{Small} & \textbf{Base} & \textbf{Large} & \textbf{Huge} \\
\midrule
100   & 38.75 & 57.24 & 61.06 & 78.86 & 82.46 \\
500   & 39.96 & 57.47 & 63.13 & 79.37 & 82.89 \\
1000  & 39.96 & 58.02 & 66.01 & 79.64 & 82.92 \\
2000  & 40.96 & 58.19 & 68.36 & 80.26 & 83.08 \\
4000  & 41.36 & 58.37 & 70.00 & 80.30 & 83.27 \\
\bottomrule
\end{tabular}
\captionof{table}{Effect of calibration set size on Top-1 accuracy under 50\% joint sparsity (both MLP and attention). DeiT-L and DeiT-H remain stable even with small calibration sets.}
\label{tab:calib_sweep}

\end{minipage}

\end{figure}

\paragraph{Ranking \& Calibration}
We compare activation-based, magnitude-based, and combined ranking strategies from Sec.~\ref{sec:ranking}, with and w/o compensation at 50\% joint sparsity. The \textbf{combined strategy} achieves the best accuracy and is used by default. Ranking ablation results appear in  Appendix \ref{app:ranking} Figure \ref{fig:ranking_ablation}.
We also study the effect of calibration size at 50\% joint pruning (Table~\ref{tab:calib_sweep}). Sensitivity decreases with model scale: larger models remain stable even with small calibration sets, while smaller models benefit more from more calibration data. We use \textbf{4000} calibration samples by default.
We use a single fixed ridge coefficient $\lambda=10^{-3}$ for every MLP and attention solve, with no per-model, per-layer, or per-sparsity tuning; on DeiT-Large at 50--70\% sparsity, accuracy varies by less than 0.06 points over $\lambda\in[10^{-5},10^{-3}]$ (Appendix~\ref{app:lambda}). Results are also stable across calibration sampling: over five independently drawn calibration sets, the sample standard deviation of Top-1 is at most 0.26 points for every pruned entry in Table~\ref{tab:corp_full}, and at most 0.06 points for DeiT-L and DeiT-H.

\paragraph{DeiT Performance}
Table~\ref{tab:corp_full} reports Top-1 accuracy, FLOPs, and parameter counts for five DeiT model sizes under 50\% sparsity. Attention pruning preserves accuracy better than MLP pruning at matched sparsity because attention contributes a smaller fraction of FLOPs ($\sim$12\% vs.\ $\sim$30\%). Accuracy preservation also improves substantially with model scale: under 50\% joint sparsity, DeiT-Huge drops only 1.7 points (84.97$\rightarrow$83.24), while DeiT-Tiny collapses to 41.69, indicating that larger ViTs show stronger redundancy in our setting for pruning with closed-form compensation.



\paragraph{Accuracy vs Sparsity}

We further explore the effect of sparsity on accuracy for DeiT-Large and DeiT-Huge.
Figure~\ref{fig:acc_sparsity} shows Top-1 accuracy as a function of sparsity for DeiT-Large and DeiT-Huge. Without compensation, accuracy collapses at high sparsity, while compensation preserves stable performance across MLP, attention, and joint pruning. The benefit increases with sparsity; for example, under 70\% joint sparsity on DeiT-Huge, compensation improves accuracy from 43.8\% to 75.4\%. DeiT-Huge remains within 2.5\% of the baseline up to 50\% sparsity across all pruning targets.
Appendix~\ref{app:rho2} directly measures the recoverability ratios $\rho^2$ of Propositions~\ref{prop:mlp-gain} and~\ref{prop:attn-gain} on DeiT-Large (50\% joint): compensation recovers 47\% (MLP) and 78\% (attention) of the pruned energy on held-out images, and the oracle-to-transfer gap stays below 0.01 (pooled), including when calibration and evaluation classes are disjoint.



\begin{table*}[t]
\centering
\footnotesize
\setlength{\tabcolsep}{3pt}
\renewcommand{\arraystretch}{1.05}

\begin{subtable}[t]{0.48\linewidth}
\centering

\begin{tabular}{lcccc}
\toprule
Method & Scope & Sparsity & Top-1 & $\Delta$ \\
\midrule
SNOWS & Attn & 2:4  & 81.01 & -3.16 \\
GRAIL & Attn & 50\% & 74.47 & -11.37 \\
\textbf{CORP} & Attn & 50\% & \textbf{81.98} & \textbf{-2.39} \\
\midrule
SNOWS & MLP & 2:4  & 69.71 & -14.46 \\
GRAIL & MLP & 50\% & 76.13 & -9.71 \\
\textbf{CORP} & MLP & 50\% & \textbf{81.80} & \textbf{-2.58} \\
\bottomrule
\end{tabular}

\caption{CORP vs.\ SNOWS and GRAIL.}
\label{tab:snows_grail}

\end{subtable}
\hfill
\begin{subtable}[t]{0.48\linewidth}
\centering

\begin{tabular}{lccc}
\toprule
Method & FLOPs$\downarrow$ & Top-1 & $\Delta$ \\
\midrule
DC-ViT & 8.3\%  & 83.78 & -0.63 \\
\textbf{CORP} & 11.6\% & 80.49 & -0.60 \\
DC-ViT & 16.6\% & 83.15 & -1.26 \\
\textbf{CORP} & 15.5\% & 79.80 & -1.29 \\
DC-ViT & 24.8\% & 81.60 & -2.81 \\
\textbf{CORP} & 23.5\% & 78.24 & -2.86 \\
\bottomrule
\end{tabular}

\caption{CORP vs.\ DC-ViT at matched FLOPs reduction.}
\label{tab:dcvit_compare}

\end{subtable}

\caption{Comparison with representative structured pruning methods.}
\label{tab:main_compare}
\end{table*}
\paragraph{Benchmark}

We compare against \textbf{VBP}~\citep{berisha2025variance}, \textbf{DC-ViT}~\citep{zhang2024dense}, \textbf{GRAIL}~\citep{tang2026grailposthoccompensationlinear}, and \textbf{SNOWS}~\citep{lucas_preserving_2024}. We exclude KCM~\citep{nova_gradient-free_2023} because its self-pretrained model is not publicly available. GRAIL results are reproduced using the official implementation under the same calibration setting as CORP; all other comparisons use reported settings from prior work and therefore differ in architecture, pruning scope, and dense-model baseline. We report relative accuracy to partially reduce this mismatch.
We compare only against methods with matching models, pruning units (channels or head dimensions), and adaptation budget; block-sparse (LPViT) and LLM-specific methods (FLAP, SliceGPT, LLM-Pruner) are discussed in Sec.~\ref{sec:related} because their sparsity is measured on a different axis.

\begin{figure}[t]
\centering
\setlength{\tabcolsep}{4pt}
\begin{minipage}[t]{0.48\linewidth}
\centering
\vspace{2pt}

\includegraphics[width=\linewidth]{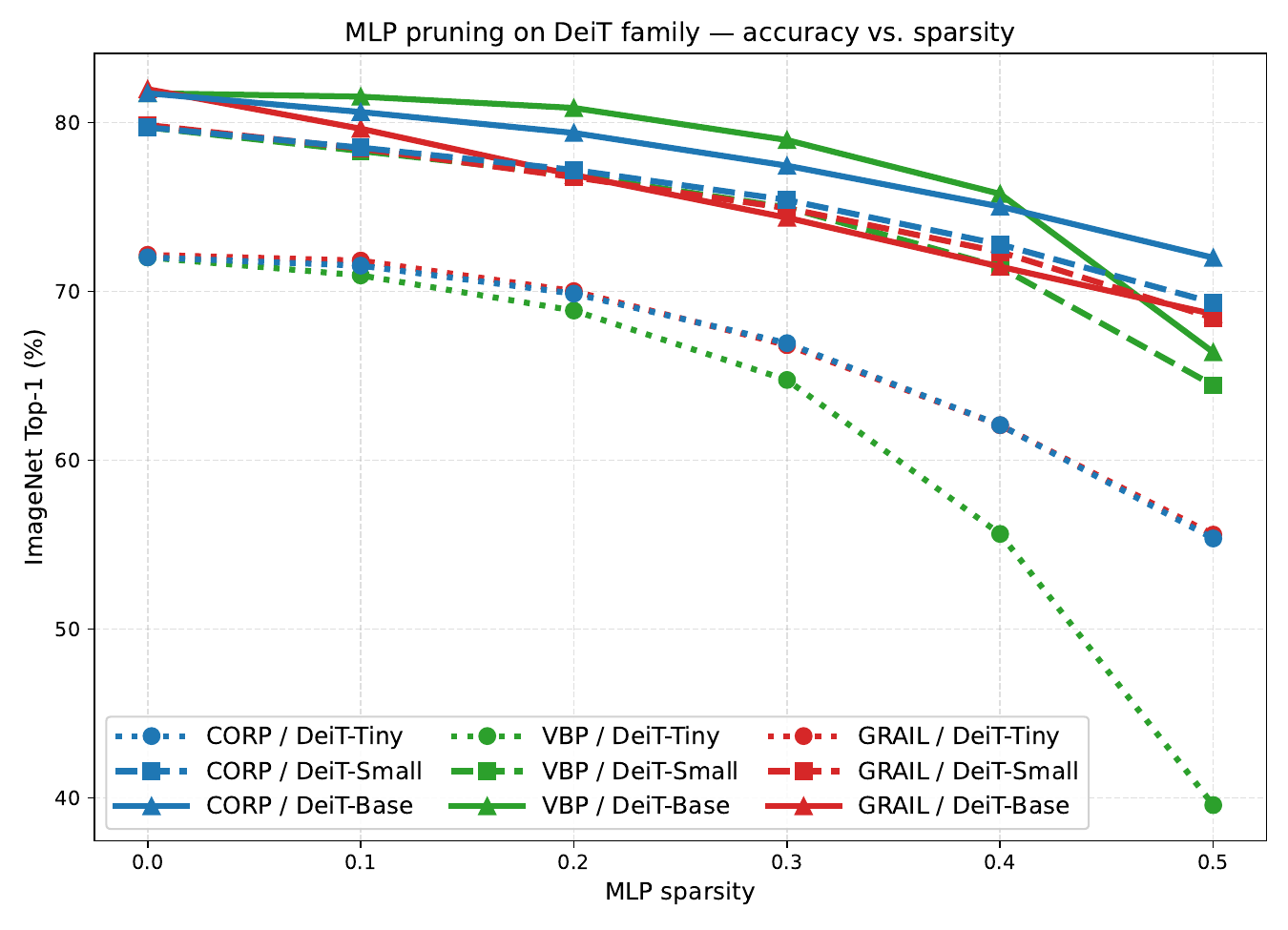}
\captionof{figure}{Comparison of CORP, VBP, and GRAIL on \textbf{MLP-only} pruning for DeiT family without fine-tuning at same MLP sparsity.}
\label{fig:deit_mlp_benchmark}

\end{minipage}
\hfill
\begin{minipage}[t]{0.48\linewidth}
\centering
\vspace{2pt}

\includegraphics[width=\linewidth]{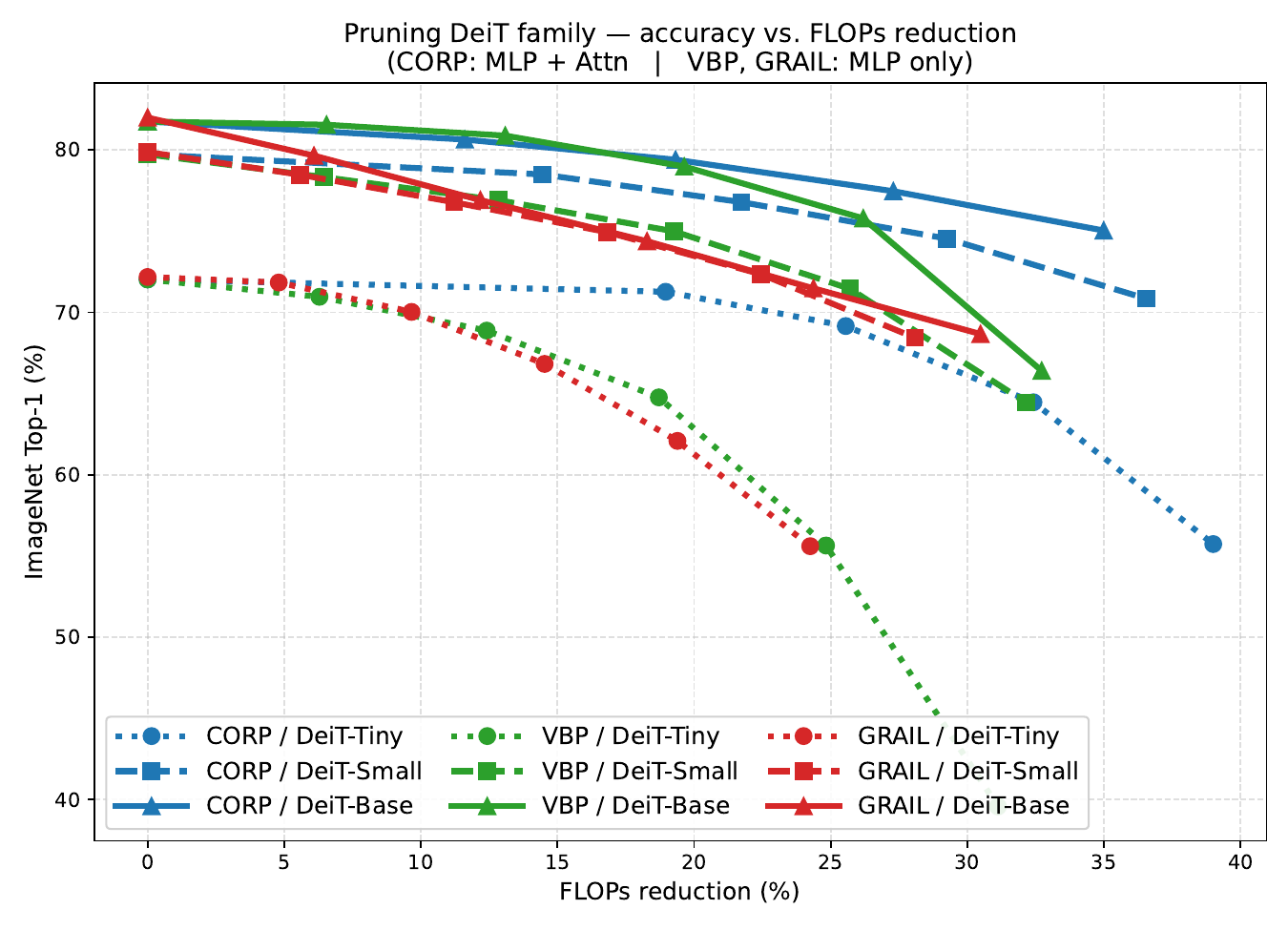}
\captionof{figure}{Pruning the DeiT family at matched FLOPs reduction. CORP prunes both MLP and attention; VBP and GRAIL prune only MLP.}
\label{fig:deit_full_benchmark}

\end{minipage}

\end{figure}

Figure~\ref{fig:deit_mlp_benchmark} compares CORP, GRAIL, and VBP under MLP-only pruning. CORP consistently outperforms VBP and remains competitive with or better than GRAIL, especially at higher sparsity. At 50\% MLP sparsity on DeiT-Base, CORP achieves 72.0\% Top-1 versus 68.6 for GRAIL and 66.4 for VBP. Figure~\ref{fig:deit_full_benchmark} compares accuracy at matched FLOPs reduction. Since VBP and GRAIL prune only MLP channels, while CORP distributes pruning across both MLP and attention, CORP achieves higher accuracy at every FLOPs target across all DeiT scales.

Table~\ref{tab:snows_grail} compares CORP with GRAIL and SNOWS on ViT-Large at $\sim$50\% compression. Under matched dimension pruning, CORP outperforms GRAIL by 7.51 points on attention (81.98 vs.\ 74.47) and 5.67 points on MLP (81.80 vs.\ 76.13), highlighting the benefit of CORP’s compensation. SNOWS uses 2:4 weight sparsity, which preserves feature dimensions and relies on sparse kernels, while CORP removes entire channels. Despite the more aggressive structured setting, CORP achieves higher accuracy on both attention (81.98 vs.\ 81.01) and MLP (81.80 vs.\ 69.71).

DC-ViT removes attention modules and prunes MLP using few-shot feature-mimicking finetuning. As shown in Table~\ref{tab:dcvit_compare} at comparable FLOPs reduction levels on ViT-B, CORP achieves nearly identical accuracy drops without finetuning. CORP therefore matches a heavily finetuned method within 0.05 $\Delta$Top-1 using only a single calibration pass without gradients. Since the dense baselines differ (84.41 for DC-ViT ViT-B vs.\ 81.74 for CORP DeiT-B), $\Delta$Top-1 is the more comparable metric.

\begin{table*}[t]
\centering
\small
\resizebox{\textwidth}{!}{
\begin{tabular}{c c c c c c c c c c}
\hline
Model & Sparsity & Top-1 & Param & FLOPs & Lat & TP & Param$\downarrow$ & FLOPs$\downarrow$ & TP$\uparrow$ \\
 &  & (\%) & (M) & (G) & (ms) & (fps) & (\%) & (\%) & (×) \\
\hline
\multirow{9}{*}{DeiT-H}
& 0.0 & 84.97 & 632.1 & 172.8 & 30.84 & 43 & 0.0 & 0.0 & 1.00 \\
& 0.1 & 84.96 & 579.7 & 153.7 & 27.70 & 48 & 8.3 & 11.1 & 1.12 \\
& 0.2 & 84.91 & 527.2 & 139.9 & 24.82 & 52 & 16.6 & 19.0 & 1.21 \\
& 0.3 & 84.66 & 474.8 & 126.2 & 22.73 & 57 & 24.9 & 27.0 & 1.32 \\
& 0.4 & 84.23 & 422.3 & 112.4 & 20.70 & 63 & 33.2 & 35.0 & 1.46 \\
& 0.5 & 83.27 & 369.9 & 98.7  & 19.16 & 71 & 41.5 & 42.9 & 1.64 \\
& 0.6 & 80.98 & 317.4 & 84.9  & 16.83 & 78 & 49.8 & 50.9 & 1.82 \\
& 0.7 & 75.38 & 265.0 & 71.2  & 14.71 & 92 & 58.1 & 58.8 & 2.13 \\
\hline
\end{tabular}}
\caption{Accuracy and efficiency trade-offs of CORP on DeiT-Huge across sparsity levels. 
We report Top-1 accuracy, parameters (M), FLOPs (G), latency (ms), and throughput (fps). 
Latency and throughput are measured with batch size 1 and 16 respectively.
Input resolution $224\times224$. 
Param$\downarrow$ and FLOPs$\downarrow$ denote reduction relative to the dense model, and TP$\uparrow$ denotes throughput speedup.}
\label{tab:efficiency_huge}
\end{table*}

\paragraph{Efficiency}
Table~\ref{tab:efficiency_huge} reports accuracy, parameter count, FLOPs, p50 latency, and throughput for DeiT-Huge across sparsity levels from 0.0 to 0.7; results for all model sizes appear in Appendix~\ref{app:eff}, Table~\ref{tab:efficiency_all}. Structured sparsity enables real GPU speedups without custom kernels. At 50\% sparsity, parameters and FLOPs drop by $\sim$40--45\%, while throughput improves by $\sim$1.5$\times$ on DeiT-Base and $\sim$1.6$\times$ on DeiT-Huge. Throughput reaches 2--2.5$\times$ at 70--80\% sparsity, though latency improves more gradually due to hardware and memory overheads. Table~\ref{tab:runtime} shows that calibration dominates runtime, while ranking and compensation add negligible overhead.

\begin{table}[t]
\centering

\begin{minipage}[t]{0.48\linewidth}
\centering
\small
\setlength{\tabcolsep}{4pt}
\renewcommand{\arraystretch}{1.0}
\begin{tabular}{lccccc}
\toprule
Model & P(M) & Cal & Rank & Comp & Tot \\
\midrule
DeiT-T & 5.7 & 81 & 0.02 & 0.40 & 81 \\
DeiT-S & 22.1 & 206 & 0.01 & 0.95 & 207 \\
DeiT-B & 86.6 & 567 & 0.02 & 3.22 & 570 \\
DeiT-L & 304.4 & 1831 & 0.05 & 13.70 & 1845 \\
DeiT-H & 632.1 & 4886 & 0.07 & 33.93 & 4920 \\
\bottomrule
\end{tabular}

\captionof{table}{\textbf{P}: parameters (M); 
\textbf{Cal}: calibration; 
\textbf{Rank}: ranking; 
\textbf{Comp}: closed-form compensation; 
\textbf{Tot}: total runtime. 
Calibration dominates the total cost, while ranking and closed-form compensation introduce negligible overhead.}
\label{tab:runtime}
\end{minipage}
\hfill
\begin{minipage}[t]{0.5\linewidth}
\centering
\small
\setlength{\tabcolsep}{5pt}
\renewcommand{\arraystretch}{1.1}
\begin{tabular}{lccc}
\toprule
Target & PPL & FLOPs (T) / $\downarrow$ & Params (B) / $\downarrow$ \\
\midrule
Baseline & 14.60 & 6.19 / 0.0\%   & 1.42 / 0.0\% \\
MLP      & 19.36 & 5.20 / 16.0\%  & 1.18 / 17.03\% \\
Attn     & 18.53 & 5.50 / 11.2\%  & 1.31 / 7.99\% \\
Both     & 25.41 & 4.51 / 27.2\%  & 1.06 / 25.01\% \\
\bottomrule
\end{tabular}
\caption{Perplexity and efficiency of CORP on OPT-1.3B at 30\% sparsity. 
\textbf{Target}: pruned component; 
\textbf{PPL}: perplexity; 
\textbf{FLOPs (T) / $\downarrow$}: FLOPs in trillions and reduction ratio; 
\textbf{Params (B) / $\downarrow$}: parameters in billions and reduction ratio.}
\label{tab:opt_efficiency}
\end{minipage}

\end{table}

\paragraph{Preliminary evidence beyond ImageNet.}


On \textbf{OPT-1.3B}, we followed \citet{frantar-sparsegpt} and use 128 calibration sequences of length 2048 from the first shard of C4 \citep{c4_dodge-etal-2021-documenting} and evaluate perplexity on WikiText-2 \citep{wiki_merity2016pointersentinelmixturemodels}.
As shown in Table \ref{tab:opt_efficiency} at 30\% sparsity, perplexity increases from 14.60 to 19.36 / 18.53 / 25.41 for MLP, attention, and joint pruning, respectively. 
The calibration--evaluation mismatch is only an initial test of robustness to distribution shift.
CORP's cost is local to each layer and head, so its computation scales with depth and width; accuracy, however, is verified only up to OPT-1.3B, and we make no claim for larger LLMs.

On \textbf{DINOv2}~\citep{oquab2024dinov2learningrobustvisual}, we evaluate CORP on NYUv2~\citep{nyuv2} depth estimation and ADE20K \citep{ADE20K} segmentation. We randomly sample 3,000 images from each dataset to form a 6,000-image unlabeled calibration set. We prune only the backbone to 50\% sparsity and keep both task heads unchanged.
As shown in Table~\ref{tab:dinov2_results}, CORP preserves downstream performance. For DINOv2-Large, RMSE changes from 0.3492 to 0.3445, while $\delta_1$ changes from 0.9466 to 0.9413. For DINOv2-Huge, RMSE improves from 0.3308 to 0.3167, while $\delta_1$ remains nearly unchanged. On segmentation, DINOv2-Huge shows only a small mIoU drop, from 0.4428 to 0.4367.
CORP thus preserves DINOv2 backbone representations on both dense prediction tasks with a mixed calibration set, even though only the shared backbone is pruned and the task heads are unchanged.

\begin{table*}[t]
\centering
\small
\setlength{\tabcolsep}{5pt}
\renewcommand{\arraystretch}{1.1}
\begin{tabular}{lcccccc}
\toprule
Model & Params (M) 
& RMSE & $\delta_1$ & mIoU \\
\midrule
Small 
& 22.1 $\rightarrow$ 14.6 
& 0.3992 $\rightarrow$ 0.4459 
& 0.9192 $\rightarrow$ 0.8750 
& 0.4198 $\rightarrow$ 0.2549 \\

Base  
& 86.6 $\rightarrow$ 56.9 
& 0.3730 $\rightarrow$ 0.4201 
& 0.9374 $\rightarrow$ 0.9118 
& 0.4368 $\rightarrow$ 0.3527 \\

Large 
& 304.4 $\rightarrow$ 198.9 
& \textbf{0.3492 $\rightarrow$ 0.3445} 
& 0.9466 $\rightarrow$ 0.9413 
& 0.4313 $\rightarrow$ 0.4011 \\

Huge  
& 1.1B $\rightarrow$ 0.74B
& \textbf{0.3308 $\rightarrow$ 0.3167} 
& 0.9529 $\rightarrow$ 0.9514 
& 0.4428 $\rightarrow$ 0.4367 \\
\bottomrule
\end{tabular}
\caption{Parameter changes and downstream performance of structured-pruned DINOv2 on depth (RMSE, $\Delta_1$) and segmentation (mIoU). Arrows denote baseline $\rightarrow$ pruned.}
\label{tab:dinov2_results}
\end{table*}

\section{Conclusion}

We presented CORP, a one-shot structured pruning method that treats pruning as layer-wise representation recovery. 
CORP uses unlabeled calibration data to fit closed-form affine and logit-space compensation, then folds the correction into the retained weights. 
Experiments on DeiT show strong preservation on larger DeiT models and measurable throughput gains, while OPT and DINOv2 results provide preliminary evidence beyond ImageNet classification. CORP trades cross-module coupling for single-pass, closed-form, training-free compensation; future work should explore layer-wise sparsity allocation and stronger calibration--test distribution shift.

\paragraph{Limitations}CORP compensates each MLP block and attention head independently from one set of calibration statistics of the unpruned model. This locality is a deliberate trade-off: it yields closed-form compensators from a single calibration pass, without labels, gradients, fine-tuning, or sequential re-calibration, and lets all solves run in parallel at negligible cost (Table~\ref{tab:runtime}). The price is that cross-module interactions are not modeled: residuals left by upstream modules are not re-fit downstream, so at high joint sparsity they can accumulate and reduce stability (Figure~\ref{fig:acc_sparsity}). The method uses a uniform sparsity ratio across layers and does not account for layer-wise redundancy differences. Because every module is solved independently, non-uniform per-layer ratios (e.g., guided by the effective ranks in Table~\ref{tab:mlp_redundancy}) can be supplied without changing the formulation; rank-aware allocation is a natural extension. CORP targets strict post-training settings with no labels, gradients, or fine-tuning, and assumes calibration data aligned with deployment. Performance may degrade under strong calibration–test shift, very high sparsity, or architectures with weaker redundancy. Finally, CORP optimizes layer-wise representation recovery rather than end-to-end task loss and thus does not provide a global optimality guarantee. Compensation is computed in float32; we do not study its composition with quantization.


\section*{Acknowledgments and Disclosure of Funding}
This work was supported in part by the Purdue
Polytechnic Institute at Purdue University.

\clearpage
\bibliographystyle{plainnat}
\bibliography{CORP}

\appendix
\onecolumn
\section{Additional Analysis of MLP Redundancy in DeiT-Base}
\label{app:mlp_redundancy}

Table~\ref{tab:mlp_redundancy} summarizes an empirical analysis of MLP activation redundancy
in \textbf{DeiT-Base}, whose MLP hidden dimension is 3072.
The calibration set consists of 4000 images randomly selected from the ImageNet training set~\citep{russakovsky2015imagenetlargescalevisual}.
Statistics are computed from calibration data collected at the output of each MLP activation layer.

First, MLP activations in DeiT-Base exhibit consistently low effective rank.
Despite a hidden dimension of 3072, the effective rank rarely exceeds 350 across all transformer blocks.
The effective rank ratio remains below 0.12 for every layer and drops below 0.06 in several cases.
This indicates that MLP activations lie in a low-dimensional subspace.

Second, activation energy is highly concentrated.
The number of channels required to explain 95\% of the activation variance ($k_{95}$)
is substantially smaller than the full feature dimension.
In early layers, fewer than 20\% of channels account for most of the activation energy.
Even in later layers, strong energy concentration persists despite increased sparsity.

Third, activation sparsity increases with network depth.
Later MLP blocks contain a large fraction of rarely active or mostly zero channels.
In some layers, activation sparsity exceeds 0.9.
These channels contribute minimally to forward computation and provide limited expressive power.

Fourth, redundancy is not uniform across layers.
Early and middle layers exhibit low effective rank with moderate sparsity,
while later layers combine low rank with extreme sparsity.
This suggests that multiple redundancy mechanisms coexist across depth in DeiT-Base.

Taken together, these observations show that DeiT-Base MLP blocks contain substantial structured redundancy.
However, directly removing channels introduces systematic bias rather than random noise.
This explains why naive structured pruning leads to severe accuracy degradation.
Effective pruning therefore requires explicit compensation to preserve the original function.

\begin{table*}[h]
\centering
\small
\setlength{\tabcolsep}{6pt}
\renewcommand{\arraystretch}{1.15}
\begin{tabular}{lcccccc}
\toprule
Layer & Dim & Eff. Rank & Rank Ratio & $k_{95}$ & $k_{95}$ Ratio & Act. Sparsity \\
\midrule
blocks.0.mlp.act  & 3072 & 164.2 & 0.053 & 542  & 0.176 & 0.23 \\
blocks.1.mlp.act  & 3072 & 136.4 & 0.044 & 336  & 0.109 & 0.30 \\
blocks.2.mlp.act  & 3072 & 294.2 & 0.096 & 1118 & 0.364 & 0.16 \\
blocks.3.mlp.act  & 3072 & 341.0 & 0.111 & 1577 & 0.513 & 0.08 \\
blocks.4.mlp.act  & 3072 & 353.8 & 0.115 & 1804 & 0.587 & 0.06 \\
blocks.5.mlp.act  & 3072 & 239.6 & 0.078 & 1783 & 0.580 & 0.07 \\
blocks.6.mlp.act  & 3072 & 279.1 & 0.091 & 1878 & 0.611 & 0.08 \\
blocks.7.mlp.act  & 3072 & 315.3 & 0.103 & 2037 & 0.663 & 0.14 \\
blocks.8.mlp.act  & 3072 & 310.5 & 0.101 & 2009 & 0.654 & 0.59 \\
blocks.9.mlp.act  & 3072 & 224.5 & 0.073 & 1617 & 0.526 & 0.86 \\
blocks.10.mlp.act & 3072 & 185.9 & 0.061 & 1203 & 0.392 & 0.92 \\
blocks.11.mlp.act & 3072 & 249.6 & 0.081 & 1646 & 0.536 & 0.19 \\
\bottomrule
\end{tabular}
\caption{Empirical analysis of MLP activation redundancy across transformer blocks in DeiT-Base.
Effective rank and $k_{95}$ quantify energy concentration in the activation covariance.
Rank ratios remain low across all layers, while activation sparsity increases in deeper blocks.}
\label{tab:mlp_redundancy}
\end{table*}

\section{Closed-Form Derivations}
\label{app:closed_form}

\subsection{MLP Affine Compensation Derivation}
\label{app:closed_form_mlp}

Let $x\in\mathbb{R}^{d}$ be the layer input and split indices into kept set $S$
and pruned set $P$:
$x=[x_S^\top\ x_P^\top]^\top$.
We fit an affine predictor
\begin{equation}
x_P \approx Bx_S + c,
\end{equation}
where $B\in\mathbb{R}^{|P|\times |S|}$ and $c\in\mathbb{R}^{|P|}$.
Let $X_S\in\mathbb{R}^{|S|\times N}$ and $X_P\in\mathbb{R}^{|P|\times N}$
stack $N$ calibration samples.
The compensated output becomes
\begin{equation}
\hat y = (W_S + W_PB) x_S +(b + W_Pc) = \widehat W_S x_S+\widehat b
\end{equation}
The ridge objective is
\begin{equation}
\min_{B,c}\ \|X_P-(BX_S+c\mathbf{1}^\top)\|_F^2+\lambda\|B\|_F^2,
\label{eq:app_ridge}
\end{equation}
with $\mathbf{1}\in\mathbb{R}^{N}$.

\paragraph{Centering.}
Let
$\mu_S=\frac{1}{N}X_S\mathbf{1}$ and $\mu_P=\frac{1}{N}X_P\mathbf{1}$.
Define centered matrices
$\bar X_S=X_S-\mu_S\mathbf{1}^\top$ and $\bar X_P=X_P-\mu_P\mathbf{1}^\top$.
Optimizing over $c$ yields
\begin{equation}
c=\mu_P-B\mu_S.
\label{eq:app_c}
\end{equation}
Substituting Eq.~\eqref{eq:app_c} into Eq.~\eqref{eq:app_ridge} reduces the problem to
\begin{equation}
\min_B\ \|\bar X_P-B\bar X_S\|_F^2+\lambda\|B\|_F^2.
\label{eq:app_ridge_centered}
\end{equation}

\paragraph{Closed form.}
Setting the gradient to zero gives
\begin{equation}
B(\bar X_S\bar X_S^\top+\lambda I)=\bar X_P\bar X_S^\top,
\end{equation}
hence
\begin{equation}
B=\bar X_P\bar X_S^\top(\bar X_S\bar X_S^\top+\lambda I)^{-1}.
\label{eq:app_B_matrix}
\end{equation}
Using second-moment notation,
\begin{equation}
\Sigma =
\begin{bmatrix}
\Sigma_{SS} & \Sigma_{SP}\\
\Sigma_{PS} & \Sigma_{PP}
\end{bmatrix},
\quad
\Sigma_{SS}=E[(x_S-\mu_S)(x_S-\mu_S)^\top],\quad
\Sigma_{PS}=E[(x_P-\mu_P)(x_S-\mu_S)^\top],
\end{equation}
the same solution can be written as
\begin{equation}
B=\Sigma_{PS}(\Sigma_{SS}+\lambda I)^{-1},\qquad
c=\mu_P-B\mu_S.
\end{equation}

\subsection{Attention Logit Compensation Derivation}
\label{app:closed_form_attn}
For clarity, we first derive the single-sample objective. In practice, CORP fits
one compensation matrix per layer and head using all calibration samples. The
calibration-set objective sums the single-sample losses, which yields a summed
linear system over samples.
We derive the Sylvester equation used in Section~\ref{sec:attn_comp}.
Let $Q=[Q_S\ Q_P]$ and $K=[K_S\ K_P]$ denote the query and key matrices, where $Q_S,K_S\in\mathbb{R}^{n\times d_h'}$.
The original logits decompose as:
\begin{equation}
    L = QK^\top = Q_S K_S^\top + Q_P K_P^\top.
\end{equation}
The missing logits after pruning are
\begin{equation}
T = Q_PK_P^\top.
\end{equation}
We approximate $T$ using the kept subspace:
\begin{equation}
T \approx Q_S M K_S^\top,
\end{equation}
and logits using only kept dimensions:
\begin{equation}
    \hat L = Q_S (I+M) K_S^\top,
\end{equation}
where $M\in\mathbb{R}^{d_h'\times d_h'}$.
We estimate $M$ by ridge regression:
\begin{equation}
\min_M\ \|T-Q_SMK_S^\top\|_F^2+\lambda\|M\|_F^2.
\label{eq:app_attn_obj}
\end{equation}

\paragraph{Normal equations.}
Let $A=Q_S$ and $B=K_S$.
Then Eq.~\eqref{eq:app_attn_obj} becomes
\begin{equation}
\min_M\ 
\|T-AM B^\top\|_F^2+\lambda\|M\|_F^2 .
\end{equation}

Define
\begin{equation}
\mathcal{L}(M)
=
\|T-AM B^\top\|_F^2+\lambda\|M\|_F^2 .
\end{equation}
Using $\|X\|_F^2=\mathrm{Tr}(X^\top X)$, we have
\begin{equation}
\mathcal{L}(M)
=
\mathrm{Tr}\left[
(T-AM B^\top)^\top(T-AM B^\top)
\right]
+
\lambda \mathrm{Tr}(M^\top M).
\end{equation}
Expanding the trace gives
\begin{equation}
\mathcal{L}(M)
=
\mathrm{Tr}(T^\top T)
-
2\mathrm{Tr}(T^\top A M B^\top)
+
\mathrm{Tr}(B M^\top A^\top A M B^\top)
+
\lambda \mathrm{Tr}(M^\top M).
\end{equation}
taking the gradient
\begin{equation}
\nabla_M \mathcal{L}(M)
=
-2A^\top T B
+
2A^\top A M B^\top B
+
2\lambda M .
\end{equation}

Setting $\nabla_M\mathcal{L}(M)=0$ gives
\begin{equation}
(A^\top A)M(B^\top B)+\lambda M=A^\top T B .
\label{eq:app_sylvester}
\end{equation}

Thus, we avoid constructing the full $n\times n$ matrix $T$.
All terms in Eq.~\eqref{eq:app_sylvester} have size $d_h'\times d_h'$.

Substituting $A=Q_S$, $B=K_S$ and $T = Q_P K_P^\top $ into Eq.~\eqref{eq:app_sylvester} gives
\begin{equation}
(Q_S^\top Q_S)M(K_S^\top K_S)+\lambda M
=
(Q_S^\top Q_P)(K_P^\top K_S).
\label{eq:app_M_matrix}
\end{equation}
Thus, we avoid constructing the full $n\times n$ matrix $T$.
All terms in Eq.~\eqref{eq:app_sylvester} have size $d_h'\times d_h'$.
This equation determines the compensation matrix $M$.

Using the identity
\begin{equation}
\mathrm{vec}\left((A^\top A)M(B^\top B)\right)
=
\left[(B^\top B)\otimes(A^\top A)\right]\mathrm{vec}(M).
\end{equation}
we obtain
\begin{equation}
\left[
(K_S^\top K_S)\otimes(Q_S^\top Q_S)
+
\lambda I
\right]\mathrm{vec}(M)
=
\mathrm{vec}\left(
(Q_S^\top Q_P)(K_P^\top K_S)
\right).
\end{equation}
Therefore,
\begin{equation}
\mathrm{vec}(M)
=
\left[
(K_S^\top K_S)\otimes(Q_S^\top Q_S)
+
\lambda I
\right]^{-1}
\mathrm{vec}\left(
(Q_S^\top Q_P)(K_P^\top K_S)
\right).
\end{equation}
Finally,
\begin{equation}
    M=\mathrm{mat}\left(\left[(K_S^\top K_S)\otimes(Q_S^\top Q_S)+\lambda I\right]^{-1} \mathrm{vec}\left((Q_S^\top Q_P)(K_P^\top K_S)\right)\right),
\end{equation}
where $\mathrm{mat}(\cdot)$ reshapes the vector into a
$d_h'\times d_h'$ matrix.

For a calibration set, the exact normal equation becomes
\begin{equation}
\sum_{b=1}^{N}
(Q_{S,b}^{\top}Q_{S,b})M(K_{S,b}^{\top}K_{S,b})
+\lambda M =\sum_{b=1}^{N}
(Q_{S,b}^{\top}Q_{P,b})(K_{P,b}^{\top}K_{S,b}).
\label{eq:attn object}
\end{equation}
Equivalently,
\begin{equation}
\left[
\sum_{b=1}^{N}
(K_{S,b}^{\top}K_{S,b})\otimes(Q_{S,b}^{\top}Q_{S,b})
+\lambda I
\right]\mathrm{vec}(M)
=
\sum_{b=1}^{N}
\mathrm{vec}\left(
(Q_{S,b}^{\top}Q_{P,b})(K_{P,b}^{\top}K_{S,b})
\right).
\label{eq:attn solution}
\end{equation}
In practice, CORP solves the calibration-summed system in Eq. \eqref{eq:attn object}\eqref{eq:attn solution}.
The single-sample Sylvester form (Eq. \eqref{eq:app_M_matrix}) is used only for derivation.
\paragraph{Folding compensation into projections.}
We seek $A,B\in\mathbb{R}^{d_h'\times d_h'}$ such that
$AB^\top\approx I+M$ and update
$\hat W_{Q,S}=W_{Q,S}A$ and $\hat W_{K,S}=W_{K,S}B$.
We use the SVD $I+M=U\Sigma V^\top$ and set
\begin{equation}
A=U\Sigma^{1/2},\quad B=V\Sigma^{1/2},
\end{equation}
which yields $AB^\top=U\Sigma V^\top=I+M$.
This preserves the reduced head dimension while compensating missing logits in expectation.

\section{Layer Distortion Bounds and the Effect of Calibration Shift}
\label{app:bounds}

\subsection{MLP}
\label{app:bounds-mlp}

We analyze the layer distortion induced by structured channel pruning under the ridge-regularized affine compensator used by CORP, and quantify how calibration--test distribution shift enters the bound. 
for a linear layer:
\begin{equation}
y = W x + b, 
\qquad W = [\, W_S \;\; W_P \,], 
\qquad x = \begin{bmatrix} x_S \\ x_P \end{bmatrix},
\label{eq:mlp-layer-app}
\end{equation}
indices are partitioned into a kept set $S$ and a pruned set $P$. After
pruning, only $x_S$ is available. Throughout this appendix, expectations are
taken under a calibration distribution $\mathcal{D}$ unless stated otherwise.
We write
\begin{equation}
\mu_S = \mathbb{E}_\mathcal{D}[x_S], \quad
\mu_P = \mathbb{E}_\mathcal{D}[x_P], \quad
\bar{x}_S = x_S - \mu_S, \quad
\bar{x}_P = x_P - \mu_P,
\end{equation}
and
\begin{equation}
\Sigma_{SS} = \mathbb{E}_\mathcal{D}[\bar{x}_S \bar{x}_S^\top], \quad
\Sigma_{PP} = \mathbb{E}_\mathcal{D}[\bar{x}_P \bar{x}_P^\top], \quad
\Sigma_{PS} = \mathbb{E}_\mathcal{D}[\bar{x}_P \bar{x}_S^\top], \quad
\Sigma_{SP} = \Sigma_{PS}^\top.
\end{equation}
The compensated parameters from Eqs.~\eqref{eq:mlp_ridge}--\eqref{eq:mlp_cov} of the main text, with ridge
regularization $\lambda \geq 0$, are
\begin{equation}
\widehat{B}_\lambda = \Sigma_{PS}\,(\Sigma_{SS} + \lambda I)^\dagger,
\qquad
\widehat{c}_\lambda = \mu_P - \widehat{B}_\lambda \mu_S,
\label{eq:mlp-ridge-pop}
\end{equation}
\begin{equation}
\widehat{W}_S = W_S + W_P \widehat{B}_\lambda,
\qquad
\widehat{b} = b + W_P \widehat{c}_\lambda,
\label{eq:mlp-fold}
\end{equation}
where $(\,\cdot\,)^\dagger$ denotes the Moore--Penrose pseudoinverse. The
layer distortion under the calibration distribution is
\begin{equation}
J_\mathcal{D}(\widehat{B}_\lambda, \widehat{c}_\lambda)
\;=\; \mathbb{E}_\mathcal{D}\!\left[\,\bigl\| Wx + b - (\widehat{W}_S x_S + \widehat{b}) \bigr\|_2^2\,\right].
\label{eq:mlp-objective}
\end{equation}

\begin{proposition}[exact MLP layer error]\label{prop:mlp-exact}
\textit{For any $\lambda \geq 0$, the ridge-compensated layer distortion
in~\eqref{eq:mlp-objective} equals}
\begin{equation}
J_\mathcal{D}(\widehat{B}_\lambda, \widehat{c}_\lambda)
\;=\; \mathrm{tr}\!\Bigl( W_P \, \Sigma_\lambda \, W_P^\top \Bigr),
\label{eq:mlp-exact}
\end{equation}
\textit{where}
\begin{equation}
\Sigma_\lambda 
\;:=\; \Sigma_{PP} 
\;-\; \Sigma_{PS}\,(\Sigma_{SS} + \lambda I)^\dagger\,\Sigma_{SP}
\;-\; \Sigma_{PS}\,\Sigma_{SP}\,(\Sigma_{SS} + \lambda I)^\dagger
\;+\; \widehat{B}_\lambda\,\Sigma_{SS}\,\widehat{B}_\lambda^\top.
\end{equation}
\textit{In the population limit $\lambda \to 0^+$ (or whenever
$\Sigma_{SS}$ is invertible), this collapses to}
\begin{equation}
J_\mathcal{D}^\star
\;=\; \mathrm{tr}\!\Bigl( W_P \, \Sigma_{P|S} \, W_P^\top \Bigr),
\qquad
\Sigma_{P|S} \;:=\; \Sigma_{PP} - \Sigma_{PS}\,\Sigma_{SS}^\dagger\,\Sigma_{SP},
\label{eq:mlp-exact-pop}
\end{equation}
\textit{the residual covariance of $x_P$ after the best linear prediction
from $x_S$.}
\end{proposition}

\textbf{Proof.} We proceed in four steps.

\emph{Step 1 (residual form).} Substituting~\eqref{eq:mlp-fold} into the
residual gives, for any $x$,
\begin{align}
W x + b - (\widehat{W}_S x_S + \widehat{b})
&= W_S x_S + W_P x_P + b 
   - \bigl[(W_S + W_P \widehat{B}_\lambda) x_S + b + W_P \widehat{c}_\lambda\bigr] \nonumber \\
&= W_P\,\bigl(x_P - \widehat{B}_\lambda x_S - \widehat{c}_\lambda\bigr).
\label{eq:mlp-residual}
\end{align}
That is, the layer residual factors through $W_P$ and the affine prediction
residual of $x_P$ from $x_S$.

\emph{Step 2 (centering the prediction residual).} Define
\begin{equation}
r \;:=\; x_P - \widehat{B}_\lambda x_S - \widehat{c}_\lambda
\;=\; \bar{x}_P - \widehat{B}_\lambda \bar{x}_S
   + \bigl(\mu_P - \widehat{B}_\lambda \mu_S - \widehat{c}_\lambda\bigr).
\end{equation}
By the definition of $\widehat{c}_\lambda$ in~\eqref{eq:mlp-ridge-pop}, the
parenthesized term vanishes, so
\begin{equation}
r \;=\; \bar{x}_P - \widehat{B}_\lambda \bar{x}_S,
\qquad \mathbb{E}_\mathcal{D}[r] = 0.
\label{eq:mlp-r}
\end{equation}

\emph{Step 3 (residual covariance).} Combining~\eqref{eq:mlp-residual}
and~\eqref{eq:mlp-r},
\begin{align}
J_\mathcal{D}(\widehat{B}_\lambda, \widehat{c}_\lambda)
&= \mathbb{E}_\mathcal{D}\!\bigl[\, \| W_P r \|_2^2 \,\bigr] 
 = \mathbb{E}_\mathcal{D}\!\bigl[\, \mathrm{tr}(W_P\, r r^\top \, W_P^\top) \,\bigr] 
 = \mathrm{tr}\!\bigl(W_P \,\mathbb{E}_\mathcal{D}[r r^\top] \, W_P^\top\bigr).
\end{align}
The covariance of $r$ expands as
\begin{align}
\mathbb{E}_\mathcal{D}[r r^\top]
&= \mathbb{E}_\mathcal{D}\!\Bigl[(\bar{x}_P - \widehat{B}_\lambda \bar{x}_S)(\bar{x}_P - \widehat{B}_\lambda \bar{x}_S)^\top\Bigr] \nonumber \\
&= \Sigma_{PP} 
   - \Sigma_{PS}\,\widehat{B}_\lambda^\top
   - \widehat{B}_\lambda\,\Sigma_{SP}
   + \widehat{B}_\lambda\,\Sigma_{SS}\,\widehat{B}_\lambda^\top.
\label{eq:mlp-Err}
\end{align}
Substituting $\widehat{B}_\lambda = \Sigma_{PS}(\Sigma_{SS}+\lambda I)^\dagger$
gives $\Sigma_\lambda$ as stated in~\eqref{eq:mlp-exact}.

\emph{Step 4 (population limit).} When $\Sigma_{SS}$ is invertible (or as
$\lambda \to 0^+$ on the column space of $\Sigma_{SS}$), the optimality
condition $\widehat{B}_0\,\Sigma_{SS} = \Sigma_{PS}$ implies
\begin{equation}
\widehat{B}_0\,\Sigma_{SS}\,\widehat{B}_0^\top 
= \Sigma_{PS}\,\Sigma_{SS}^\dagger\,\Sigma_{SP}
= \widehat{B}_0\,\Sigma_{SP}
= \Sigma_{PS}\,\widehat{B}_0^\top,
\end{equation}
so~\eqref{eq:mlp-Err} simplifies to
\begin{equation}
\mathbb{E}_\mathcal{D}[r r^\top]
= \Sigma_{PP} - \Sigma_{PS}\,\Sigma_{SS}^\dagger\,\Sigma_{SP}
= \Sigma_{P|S},
\end{equation}
which is the Schur complement of $\Sigma$ on the kept block, yielding
\eqref{eq:mlp-exact-pop}. \hfill$\square$

\paragraph{Interpretation.}
The minimum layer distortion equals the trace of the Schur complement
$\Sigma_{P|S}$ weighted by the removed columns $W_P$. Pruning is loss-free
in the affine family iff $W_P\,\Sigma_{P|S}^{1/2} = 0$, i.e., the removed
columns lie in the kernel of the residual covariance of $x_P$ given $x_S$.
Statistically, $\Sigma_{P|S}$ is the conditional covariance of $x_P$ given a
linear function of $x_S$; the result does not require $x$ itself to be
low-rank, only that the pruned activations be linearly predictable from the
kept ones in the directions selected by $W_P$.

\begin{proposition}[MLP compensation gain]\label{prop:mlp-gain}
\textit{Define the uncompensated layer distortion}
\begin{equation}
J_{\mathrm{uncomp}} 
\;:=\; \mathbb{E}_\mathcal{D}\!\left[\,\bigl\| Wx + b - (W_S x_S + b) \bigr\|_2^2\,\right]
\;=\; \mathrm{tr}\!\bigl(W_P\,\Sigma_{PP}\,W_P^\top\bigr)
   + \bigl\| W_P\,\mu_P \bigr\|_2^2.
\end{equation}
\textit{Then in the population limit ($\lambda\to 0^+$), the compensation gain is}
\begin{equation}
J_{\mathrm{uncomp}} - J_\mathcal{D}^\star
\;=\; \underbrace{\mathrm{tr}\!\bigl(W_P\,\Sigma_{PS}\,\Sigma_{SS}^\dagger\,\Sigma_{SP}\,W_P^\top\bigr)}_{\text{variance-explained term} \,\geq\, 0}
\;+\; \underbrace{\bigl\| W_P\,\mu_P \bigr\|_2^2}_{\text{bias-correction term} \,\geq\, 0},
\label{eq:mlp-gain}
\end{equation}
\textit{with equality (zero gain) iff $W_P\,\mu_P = 0$ and $W_P\,\Sigma_{PS} = 0$.
Equivalently, defining}
\begin{equation}
\rho^2_{W_P}
\;:=\; \frac{\mathrm{tr}\!\bigl(W_P\,\Sigma_{PS}\,\Sigma_{SS}^\dagger\,\Sigma_{SP}\,W_P^\top\bigr)}
            {\mathrm{tr}\!\bigl(W_P\,\Sigma_{PP}\,W_P^\top\bigr)}
\;\in\; [0,1],
\label{eq:mlp-rho}
\end{equation}
\textit{the centered ratio satisfies $J_\mathcal{D}^\star\big/\bigl(J_{\mathrm{uncomp}} - \|W_P\mu_P\|_2^2\bigr) = 1 - \rho^2_{W_P}$.}
\end{proposition}

\textbf{Proof.} We proceed in three steps.

\emph{Step 1 (uncompensated error).} The uncompensated residual is
$Wx + b - (W_S x_S + b) = W_P x_P$. Hence
\begin{align}
J_{\mathrm{uncomp}}
&= \mathbb{E}_\mathcal{D}\!\bigl[\, \| W_P x_P \|_2^2 \,\bigr]
= \mathrm{tr}\!\Bigl(W_P\,\mathbb{E}_\mathcal{D}[x_P x_P^\top]\,W_P^\top\Bigr).
\end{align}
Using $\mathbb{E}[x_P x_P^\top] = \Sigma_{PP} + \mu_P \mu_P^\top$,
\begin{align}
J_{\mathrm{uncomp}}
&= \mathrm{tr}\!\bigl(W_P \Sigma_{PP} W_P^\top\bigr) 
 + \mathrm{tr}\!\bigl(W_P \mu_P \mu_P^\top W_P^\top\bigr) \nonumber \\
&= \mathrm{tr}\!\bigl(W_P \Sigma_{PP} W_P^\top\bigr) + \| W_P \mu_P \|_2^2.
\label{eq:mlp-Junc}
\end{align}

\emph{Step 2 (subtract the compensated optimum).}
By Proposition~\ref{prop:mlp-exact} in the population limit,
$J_\mathcal{D}^\star = \mathrm{tr}\!\bigl(W_P\,\Sigma_{PP}\,W_P^\top\bigr)
 - \mathrm{tr}\!\bigl(W_P\,\Sigma_{PS}\,\Sigma_{SS}^\dagger\,\Sigma_{SP}\,W_P^\top\bigr)$.
Subtracting from~\eqref{eq:mlp-Junc},
\begin{equation}
J_{\mathrm{uncomp}} - J_\mathcal{D}^\star
= \mathrm{tr}\!\bigl(W_P\,\Sigma_{PS}\,\Sigma_{SS}^\dagger\,\Sigma_{SP}\,W_P^\top\bigr)
 + \| W_P \mu_P \|_2^2.
\end{equation}

\emph{Step 3 (non-negativity).} The matrix
$\Sigma_{PS}\,\Sigma_{SS}^\dagger\,\Sigma_{SP}$ is positive semidefinite:
writing $M := \Sigma_{SS}^{\dagger/2}\,\Sigma_{SP}$ (well-defined on
$\mathrm{range}(\Sigma_{SS})$),
\begin{equation}
\Sigma_{PS}\,\Sigma_{SS}^\dagger\,\Sigma_{SP} = M^\top M \,\succeq\, 0,
\end{equation}
so its trace weighted by $W_P\,W_P^\top \succeq 0$ is non-negative. The
bias-correction term is a squared Euclidean norm and is also non-negative.
Equality of either term to zero requires the corresponding factor to
vanish, giving the stated equality condition. The ratio form follows by
dividing the centered identity
$\mathrm{tr}(W_P \Sigma_{PP} W_P^\top) - J_\mathcal{D}^\star
= \mathrm{tr}(W_P \Sigma_{PS} \Sigma_{SS}^\dagger \Sigma_{SP} W_P^\top)$
by $\mathrm{tr}(W_P \Sigma_{PP} W_P^\top)$.
\hfill$\square$

\paragraph{Interpretation.}
Compensation strictly reduces the layer error whenever either (i)~$W_P\mu_P \neq 0$
(pruned channels have non-zero mean and the bias correction in
$\widehat{c}_\lambda$ matters, e.g., post-GELU MLPs), or (ii)~the kept
activations carry any linear signal about the pruned ones in the directions
of $W_P$. The quantity $\rho^2_{W_P}$ is a multivariate coefficient of
determination: it measures the fraction of the variance of $W_P x_P$ that
is linearly explained by $x_S$. The bias and variance terms in
\eqref{eq:mlp-gain} correspond exactly to the two parts of the closed-form
solution in~\eqref{eq:mlp-ridge-pop}: $\widehat{c}_\lambda$ removes the
mean term, and $\widehat{B}_\lambda$ captures the predictable variance.
Both $\Sigma_{PS}$ and $\Sigma_{SS}$ are already computed during
compensation (Algorithm~\ref{alg:comp_mlp}), so $\rho^2_{W_P}$ is available at no additional
cost; measured values are reported in Appendix~\ref{app:rho2}.

\begin{proposition}[calibration--test shift, sketch]\label{prop:mlp-shift}
\textit{Let $\widehat{B}_\lambda, \widehat{c}_\lambda$ be the ridge
estimators in~\eqref{eq:mlp-ridge-pop} fitted from $N$ i.i.d.\ calibration
samples $\{x^{(i)}\}_{i=1}^N$ drawn from $\mathcal{D}$, and let
$(B_t^\star, c_t^\star)$ be the population-optimal affine predictor under a
deployment distribution $\mathcal{D}_t$. Define the test-time layer error}
\begin{equation}
\mathcal{E}_t(\widehat{B}_\lambda, \widehat{c}_\lambda)
\;:=\; \mathbb{E}_{\mathcal{D}_t}\!\left[\,\bigl\| W_P\,(x_P - \widehat{B}_\lambda x_S - \widehat{c}_\lambda) \bigr\|_2^2\,\right].
\end{equation}
\textit{Assume (A1) $\| x \|_2 \leq R$ almost surely under both $\mathcal{D}$
and $\mathcal{D}_t$, and (A2) $\Sigma_{SS}^{(c)} + \lambda I \succeq \sigma_{\min} I$
for some $\sigma_{\min} > 0$. Then with probability at least $1-\delta$
over the calibration draw,}
\begin{equation}
\mathcal{E}_t(\widehat{B}_\lambda, \widehat{c}_\lambda)
\;\leq\;
2\,J_{\mathcal{D}_t}^\star
\;+\; C_1\, \| W_P \|_F^2 \,\frac{R^4\, |S|}{\sigma_{\min}^2\, N}\,\log\!\frac{|S|}{\delta}
\;+\; C_2\, \| W_P \|_F^2 \,\Bigl( \bigl\| \Sigma^{(c)} - \Sigma^{(t)} \bigr\|_{\mathrm{op}}
   + \bigl\| \mu^{(c)} - \mu^{(t)} \bigr\|_2^2 \Bigr),
\label{eq:mlp-shift}
\end{equation}
\textit{for absolute constants $C_1, C_2 > 0$, where $J_{\mathcal{D}_t}^\star$
is the irreducible affine residual on $\mathcal{D}_t$ given by
Proposition~\ref{prop:mlp-exact} applied to $\mathcal{D}_t$.}
\end{proposition}

\textbf{Proof sketch.} The argument follows three standard steps.

\emph{(i) Bias--variance split.} Using the decomposition
\begin{equation}
x_P - \widehat{B}_\lambda x_S - \widehat{c}_\lambda
= \bigl(x_P - B_t^\star x_S - c_t^\star\bigr)
+ \bigl(B_t^\star - \widehat{B}_\lambda\bigr) x_S
+ \bigl(c_t^\star - \widehat{c}_\lambda\bigr),
\end{equation}
and the inequality $\|u + v\|_2^2 \leq 2\|u\|_2^2 + 2\|v\|_2^2$ to obtain
\begin{equation}
\mathcal{E}_t(\widehat{B}_\lambda, \widehat{c}_\lambda)
\leq 2\,J_{\mathcal{D}_t}^\star
   + 2\,\|W_P\|_F^2\, \mathbb{E}_{\mathcal{D}_t}\!\bigl[\,\bigl\|(B_t^\star - \widehat{B}_\lambda)x_S + (c_t^\star - \widehat{c}_\lambda)\bigr\|_2^2\,\bigr].
\end{equation}

\emph{(ii) Estimation error on $\mathcal{D}$.} Decompose
$B_t^\star - \widehat{B}_\lambda = (B_t^\star - B_c^\star) + (B_c^\star - \widehat{B}_\lambda)$
where $B_c^\star$ is the population ridge solution on $\mathcal{D}$.
Under (A1)--(A2), matrix Bernstein concentration applied to
$\widehat{\Sigma}_{SS}^{(c)}$ and $\widehat{\Sigma}_{PS}^{(c)}$
\citep{tropp2015introduction} yields
\begin{equation}
\bigl\| B_c^\star - \widehat{B}_\lambda \bigr\|_F^2
\;\leq\; C_1' \,\frac{R^4\, |S|}{\sigma_{\min}^2\, N}\,\log\!\frac{|S|}{\delta},
\end{equation}
with probability at least $1-\delta$, and an analogous Hoeffding bound for
$\mu^{(c)}$ controls $\|c_c^\star - \widehat{c}_\lambda\|_2^2$ at the same
rate. The fast-rate variant of \citet{hsu2014random} gives the explicit
constants under sub-Gaussian relaxations of (A1).

\emph{(iii) Distribution shift on $\mathcal{D}_t$.} The remaining piece
$B_t^\star - B_c^\star$ is bounded via standard matrix perturbation
\citep[e.g.,][Cor.~6.3.8]{horn2013matrix}: a $C^1$ perturbation of
$\Sigma_{SS}$ propagates through the (pseudo)inverse, yielding
\begin{equation}
\bigl\| B_t^\star - B_c^\star \bigr\|_F^2
\;\leq\; C_2'\, \bigl\| \Sigma^{(c)} - \Sigma^{(t)} \bigr\|_{\mathrm{op}}^2 \big/ \sigma_{\min}^2,
\end{equation}
and a triangle inequality on the means controls the bias drift.

Combining (i)--(iii) and absorbing constants into $C_1, C_2$ gives
\eqref{eq:mlp-shift}.
\hfill$\square$

\paragraph{Interpretation.}
The test-time error decomposes into three terms. The first,
$2\,J_{\mathcal{D}_t}^\star$, is the irreducible bilinear residual on the
deployment distribution and is the same Schur-complement quantity from
Proposition~\ref{prop:mlp-exact} applied to $\mathcal{D}_t$; it is the floor any
distribution-conditional method must pay. The second term decays as
$|S|/N$ and gives the calibration sample-complexity guidance: doubling
$|S|$ (less aggressive pruning, larger kept set) doubles the calibration
budget needed for the same estimation error, and the dependence on
$\sigma_{\min}^{-2}$ explains why aggressive pruning that leaves a poorly
conditioned $\Sigma_{SS}$ inflates the bound. The third term grows
linearly in the operator-norm gap between calibration and deployment
covariances, providing a controlled (rather than catastrophic) sensitivity
to mismatch and supporting the empirical robustness observed in our
OPT--C4/WikiText-2 experiment (Section~\ref{sec:exp}).

\subsection{Attention QK}
\label{app:bounds-attn}

We now derive the analogous bounds for the attention compensation. Following
the notation of Section \ref{sec:attn_comp}, fix a layer and a head; let $Q, K \in \mathbb{R}^{n \times d_h'}$
denote per-sample query and key matrices restricted to that head, and partition
$Q = [\,Q_S \;\; Q_P\,]$, $K = [\,K_S \;\; K_P\,]$ along the head dimension into
kept indices $S$ and pruned indices $P$. The original logits decompose as
\begin{equation}
L \;=\; Q K^\top \;=\; Q_S K_S^\top \;+\; \underbrace{Q_P K_P^\top}_{=:\, T},
\label{eq:attn-decomp}
\end{equation}
and the missing piece $T = Q_P K_P^\top$ is what compensation tries to recover
through a small matrix $M \in \mathbb{R}^{d_h' \times d_h'}$.

Let $\mathcal{D} = \{x^{(b)}\}_{b=1}^N$ be the calibration set, and write
$Q_{S,b}, K_{S,b}, Q_{P,b}, K_{P,b}, T_b$ for the per-sample matrices. The
compensation objective from Eqs.~\eqref{eq:attn_ridge_obj} and~\eqref{eq:attn-ridge1} of the main text is the empirical
ridge problem
\begin{equation}
\widehat{M}_\lambda
\;:=\; \arg\min_{M}\;
\sum_{b=1}^{N} \bigl\| T_b - Q_{S,b}\,M\,K_{S,b}^\top \bigr\|_F^2
+ \lambda\, \| M \|_F^2.
\label{eq:attn-objective}
\end{equation}
Vectorizing $M$ via $m := \mathrm{vec}(M)$ and using the Kronecker identity
$\mathrm{vec}(Q_S\,M\,K_S^\top) = (K_S \otimes Q_S)\,\mathrm{vec}(M)$, define
the empirical Gram tensor and cross-product
\begin{equation}
G \;:=\; \sum_{b=1}^{N} (K_{S,b}^\top K_{S,b}) \otimes (Q_{S,b}^\top Q_{S,b})
\;\in\; \mathbb{R}^{d_h'^2 \times d_h'^2},
\label{eq:attn-G}
\end{equation}
\begin{equation}
h \;:=\; \sum_{b=1}^{N} \mathrm{vec}\!\Bigl( (Q_{S,b}^\top Q_{P,b})(K_{P,b}^\top K_{S,b}) \Bigr)
\;\in\; \mathbb{R}^{d_h'^2}.
\label{eq:attn-h}
\end{equation}
The closed-form ridge solution is
\begin{equation}
\mathrm{vec}(\widehat{M}_\lambda) \;=\; (G + \lambda I)^\dagger\, h.
\label{eq:attn-ridge}
\end{equation}

\begin{proposition}[exact attention layer error]\label{prop:attn-exact}
\textit{For any $\lambda \geq 0$, the ridge-compensated logit distortion satisfies}
\begin{equation}
J_\mathcal{D}(\widehat{M}_\lambda)
\;:=\; \sum_{b=1}^{N} \bigl\| T_b - Q_{S,b}\, \widehat{M}_\lambda\, K_{S,b}^\top \bigr\|_F^2
\;=\; \sum_{b=1}^{N}\!\bigl\| T_b \bigr\|_F^2
\;-\; 2\, h^\top (G+\lambda I)^\dagger\, h
\;+\; h^\top (G+\lambda I)^\dagger\, G\, (G+\lambda I)^\dagger\, h.
\label{eq:attn-exact-finite}
\end{equation}
\textit{In the population limit $\lambda \to 0^+$ (or whenever $G$ is invertible),
this collapses to}
\begin{equation}
J_\mathcal{D}^\star
\;=\; \sum_{b=1}^{N}\!\bigl\| T_b \bigr\|_F^2 \;-\; h^\top G^\dagger\, h.
\label{eq:attn-exact-pop}
\end{equation}
\end{proposition}

\textbf{Proof.} We proceed in four steps.

\emph{Step 1 (vectorize the objective).} Let $m = \mathrm{vec}(M)$ and
$t_b = \mathrm{vec}(T_b)$. The Kronecker identity
$\mathrm{vec}(AMB^\top) = (B \otimes A)\,\mathrm{vec}(M)$ applied to
$A = Q_{S,b}$, $B = K_{S,b}$ gives
\begin{equation}
\mathrm{vec}(Q_{S,b}\,M\,K_{S,b}^\top) \;=\; \Phi_b\,m,
\qquad
\Phi_b \;:=\; K_{S,b} \otimes Q_{S,b} \;\in\; \mathbb{R}^{n^2 \times d_h'^2}.
\end{equation}
Hence the per-sample loss equals $\| T_b - Q_{S,b}\,M\,K_{S,b}^\top \|_F^2 = \| t_b - \Phi_b\,m \|_2^2$, and the empirical objective in~\eqref{eq:attn-objective} becomes
\begin{equation}
J_\mathcal{D}(M) + \lambda\| m \|_2^2
\;=\; \sum_{b=1}^{N} \| t_b - \Phi_b\,m \|_2^2 \;+\; \lambda\, \| m \|_2^2.
\label{eq:attn-obj-vec}
\end{equation}

\emph{Step 2 (identify $G$ and $h$).} Expanding the squared norm,
\begin{equation}
\sum_{b=1}^{N} \| t_b - \Phi_b\,m \|_2^2
\;=\; \sum_{b=1}^{N} \| t_b \|_2^2
\;-\; 2\, m^\top \!\sum_{b=1}^{N}\Phi_b^\top t_b
\;+\; m^\top\! \biggl(\sum_{b=1}^{N}\Phi_b^\top\Phi_b\biggr)\!m.
\end{equation}
Two algebraic identities reduce this to the form involving $G$ and $h$. First,
\begin{equation}
\Phi_b^\top \Phi_b 
\;=\; (K_{S,b} \otimes Q_{S,b})^\top (K_{S,b} \otimes Q_{S,b})
\;=\; (K_{S,b}^\top K_{S,b}) \otimes (Q_{S,b}^\top Q_{S,b}),
\end{equation}
so $\sum_b \Phi_b^\top \Phi_b = G$. Second, using
$(B \otimes A)^\top \mathrm{vec}(C) = \mathrm{vec}(A^\top C\, B)$ with
$A = Q_{S,b}$, $B = K_{S,b}$, $C = T_b = Q_{P,b} K_{P,b}^\top$,
\begin{equation}
\Phi_b^\top t_b 
\;=\; \mathrm{vec}\!\bigl( Q_{S,b}^\top\, Q_{P,b} K_{P,b}^\top\, K_{S,b} \bigr)
\;=\; \mathrm{vec}\!\bigl( (Q_{S,b}^\top Q_{P,b})(K_{P,b}^\top K_{S,b}) \bigr),
\end{equation}
so $\sum_b \Phi_b^\top t_b = h$. Substituting into~\eqref{eq:attn-obj-vec}:
\begin{equation}
J_\mathcal{D}(M) + \lambda \|m\|_2^2
\;=\; \sum_{b=1}^{N} \| t_b \|_2^2 \;-\; 2\, m^\top h \;+\; m^\top (G + \lambda I)\, m.
\label{eq:attn-obj-quad}
\end{equation}

\emph{Step 3 (substitute the ridge optimum).} Plugging
$\widehat{m}_\lambda = (G+\lambda I)^\dagger h$ into~\eqref{eq:attn-obj-quad}
and dropping the regularizer term to recover $J_\mathcal{D}$ alone,
\begin{align}
J_\mathcal{D}(\widehat{M}_\lambda)
&= \sum_{b=1}^{N} \| t_b \|_2^2 \;-\; 2\, h^\top (G+\lambda I)^\dagger h
   \;+\; h^\top (G+\lambda I)^\dagger\, G\, (G+\lambda I)^\dagger\, h,
\end{align}
which is~\eqref{eq:attn-exact-finite}. Note that
$\sum_b \| t_b \|_2^2 = \sum_b \| T_b \|_F^2$ by the Frobenius--$\ell_2$ identity for $\mathrm{vec}(\cdot)$.

\emph{Step 4 (population limit).} When $\lambda \to 0^+$ on the column space of $G$,
$(G+\lambda I)^\dagger \to G^\dagger$, and the optimality condition
$G\, G^\dagger\, h = h$ (since $h \in \mathrm{range}(G)$ by construction) gives
\begin{equation}
h^\top G^\dagger\, G\, G^\dagger\, h \;=\; h^\top G^\dagger\, h,
\end{equation}
so the last two terms in~\eqref{eq:attn-exact-finite} combine to $-h^\top G^\dagger h$,
yielding~\eqref{eq:attn-exact-pop}. \hfill$\square$

\paragraph{Interpretation.}
The minimum logit distortion equals the total squared norm of the missing
logits $\sum_b \| T_b \|_F^2$ minus a non-negative quantity $h^\top G^\dagger h$
that quantifies how much of $T_b$ is reconstructible from the kept Kronecker
subspace $\mathrm{span}(K_{S,b} \otimes Q_{S,b})$. The matrix $G$ and vector
$h$ are exactly the LHS and RHS of the calibration-summed normal
equation~\eqref{eq:attn object} (Appendix~\ref{app:closed_form_attn}) and are computed during compensation, so
this exact error is available at no additional cost.

\begin{proposition}[attention compensation gain]\label{prop:attn-gain}
\textit{Define the uncompensated logit distortion}
\begin{equation}
J_{\mathrm{uncomp}}^{\,\mathrm{attn}}
\;:=\; \sum_{b=1}^{N} \bigl\| Q_P\, K_P^\top \bigr\|_F^2 \big|_{x = x^{(b)}}
\;=\; \sum_{b=1}^{N} \| T_b \|_F^2,
\end{equation}
\textit{obtained by setting $M = 0$ in~\eqref{eq:attn-objective}. Then in the
population limit ($\lambda \to 0^+$) the compensation gain is}
\begin{equation}
J_{\mathrm{uncomp}}^{\,\mathrm{attn}} \;-\; J_\mathcal{D}^\star
\;=\; h^\top G^\dagger\, h \;\geq\; 0,
\label{eq:attn-gain}
\end{equation}
\textit{with equality (zero gain) iff $h = 0$, i.e., the kept Q/K subspace is
orthogonal to the missing logits in the bilinear sense
$Q_{S,b}^\top Q_{P,b} K_{P,b}^\top K_{S,b} = 0$ for almost every $b$.
Equivalently, defining the bilinear coefficient of determination}
\begin{equation}
\rho^2_{\mathrm{attn}}
\;:=\; \frac{h^\top G^\dagger\, h}{\sum_{b=1}^{N} \| T_b \|_F^2}
\;\in\; [0,1],
\label{eq:attn-rho}
\end{equation}
\textit{the ratio satisfies $J_\mathcal{D}^\star \big/ J_{\mathrm{uncomp}}^{\,\mathrm{attn}} = 1 - \rho^2_{\mathrm{attn}}$.}
\end{proposition}

\textbf{Proof.} We proceed in two steps.

\emph{Step 1 (uncompensated error).} Setting $m = 0$ in~\eqref{eq:attn-obj-quad}
gives
\begin{equation}
J_\mathcal{D}(M{=}0) \;=\; \sum_{b=1}^{N} \| t_b \|_2^2 \;=\; \sum_{b=1}^{N} \| T_b \|_F^2 \;=\; J_{\mathrm{uncomp}}^{\,\mathrm{attn}}.
\end{equation}

\emph{Step 2 (subtract the compensated optimum).} By Proposition~\ref{prop:attn-exact}
in the population limit~\eqref{eq:attn-exact-pop},
$J_\mathcal{D}^\star = \sum_b \| T_b \|_F^2 - h^\top G^\dagger h$. Subtracting,
\begin{equation}
J_{\mathrm{uncomp}}^{\,\mathrm{attn}} - J_\mathcal{D}^\star \;=\; h^\top G^\dagger\, h.
\end{equation}
The matrix $G$ is positive semidefinite as a sum of Kronecker products of
PSD matrices,
\begin{equation}
G \;=\; \sum_{b=1}^{N} (K_{S,b}^\top K_{S,b}) \otimes (Q_{S,b}^\top Q_{S,b}) \;\succeq\; 0,
\end{equation}
so $G^\dagger$ is also PSD on $\mathrm{range}(G)$, and $h \in \mathrm{range}(G)$
by construction (each summand of $h$ lies in the same column span as the
corresponding summand of $G$). Hence $h^\top G^\dagger h \geq 0$, with equality
iff $h = 0$. Dividing by $J_{\mathrm{uncomp}}^{\,\mathrm{attn}}$ yields the
ratio form~\eqref{eq:attn-rho}. \hfill$\square$

\paragraph{Interpretation.}
$\rho^2_{\mathrm{attn}}$ is the attention analogue of the regression
$R^2$ in Proposition~\ref{prop:mlp-gain}: it measures the fraction of the squared
Frobenius norm of the missing logits $T_b = Q_{P,b} K_{P,b}^\top$ that lies
in the bilinear span of $Q_{S,b}\, M\, K_{S,b}^\top$ over $M$. Compensation
strictly improves on uncompensated pruning whenever $\rho^2_{\mathrm{attn}} > 0$,
i.e., whenever the kept Q/K subspaces carry any bilinear signal about the
removed ones. Both $G$ and $h$ are formed during compensation, so per-layer
or per-head $\rho^2_{\mathrm{attn}}$ values can be reported as a diagnostic
without any additional computation.

\begin{proposition}[attention calibration--test shift, sketch]\label{prop:attn-shift}
\textit{Let $\widehat{M}_\lambda$ be the ridge solution
in~\eqref{eq:attn-ridge} fitted from $N$ i.i.d.\ calibration samples drawn
from $\mathcal{D}$. Let $M_t^\star$ be the population-optimal compensator
on a deployment distribution $\mathcal{D}_t$, and define the test-time
distortion}
\begin{equation}
\mathcal{E}_t(\widehat{M}_\lambda)
\;:=\; \mathbb{E}_{\mathcal{D}_t}\!\bigl[\, \| Q_P K_P^\top - Q_S\,\widehat{M}_\lambda\, K_S^\top \|_F^2 \,\bigr].
\end{equation}
\textit{Assume (B1) $\| q_j \|_2,\, \| k_j \|_2 \leq R$ almost surely
under both $\mathcal{D}$ and $\mathcal{D}_t$ for every column index $j$,
and (B2) $\mathbb{E}_{\mathcal{D}}[G/N] + \lambda I \succeq \sigma_{\min} I$
for some $\sigma_{\min} > 0$, where $G$ is normalized by sample count.
Then with probability at least $1 - \delta$ over the calibration draw,}
\begin{equation}
\mathcal{E}_t(\widehat{M}_\lambda)
\;\leq\;
2\, J_{\mathcal{D}_t}^\star
\;+\; C_3\, \frac{R^4\, d_h'^2}{\sigma_{\min}^2\, N}\,\log\!\frac{d_h'}{\delta}
\;+\; C_4\, \Bigl( \bigl\| G^{(c)}/N - G^{(t)} \bigr\|_{\mathrm{op}}
   + \bigl\| h^{(c)}/N - h^{(t)} \bigr\|_2^2 \Bigr),
\label{eq:attn-shift}
\end{equation}
\textit{for absolute constants $C_3, C_4 > 0$, where $J_{\mathcal{D}_t}^\star$
is the irreducible bilinear residual on $\mathcal{D}_t$ given by
Proposition~\ref{prop:attn-exact} applied to $\mathcal{D}_t$, and $G^{(t)}$, $h^{(t)}$ are
the population versions on $\mathcal{D}_t$.}
\end{proposition}

\textbf{Proof sketch.} Three standard steps, in direct analogy to
Proposition~\ref{prop:mlp-shift}.

\emph{(i) Bias--variance split.} Add and subtract the test-optimal
compensator $M_t^\star$ inside the residual:
\begin{equation}
Q_P K_P^\top - Q_S\,\widehat{M}_\lambda\, K_S^\top
\;=\; \bigl( Q_P K_P^\top - Q_S\,M_t^\star\, K_S^\top \bigr)
\;+\; Q_S\,(M_t^\star - \widehat{M}_\lambda)\, K_S^\top.
\end{equation}
Applying $\| A + B \|_F^2 \leq 2 \| A \|_F^2 + 2 \| B \|_F^2$ and taking
expectations under $\mathcal{D}_t$ gives
\begin{equation}
\mathcal{E}_t(\widehat{M}_\lambda)
\leq 2\, J_{\mathcal{D}_t}^\star
   + 2\, \mathbb{E}_{\mathcal{D}_t}\!\bigl[\, \| Q_S\,(M_t^\star - \widehat{M}_\lambda)\, K_S^\top \|_F^2 \,\bigr].
\end{equation}

\emph{(ii) Estimation error on $\mathcal{D}$.} Decompose
$M_t^\star - \widehat{M}_\lambda = (M_t^\star - M_c^\star) + (M_c^\star - \widehat{M}_\lambda)$
where $M_c^\star$ is the population ridge solution on $\mathcal{D}$.
Under (B1)--(B2), matrix Bernstein concentration applied to the empirical
versions of $G$ and $h$ \citep{tropp2015introduction} yields
\begin{equation}
\bigl\| \mathrm{vec}(M_c^\star - \widehat{M}_\lambda) \bigr\|_2^2
\;\leq\; C_3'\, \frac{R^4\, d_h'^2}{\sigma_{\min}^2\, N}\, \log\!\frac{d_h'}{\delta},
\end{equation}
with probability at least $1 - \delta$. The dimension of $\mathrm{vec}(M)$
is $d_h'^2$, which appears in both the dimension factor of Bernstein and the
log term. The fast-rate analysis of \citet{hsu2014random}, transferred to
the bilinear ridge setting, gives the explicit constants under sub-Gaussian
relaxations of (B1).

\emph{(iii) Distribution shift.} Bound $M_t^\star - M_c^\star$ via
pseudoinverse perturbation \citep[e.g.,][]{stewart1977perturbation}: a
$C^1$ perturbation of $G$ propagates through $G^\dagger$, giving
\begin{equation}
\bigl\| \mathrm{vec}(M_t^\star - M_c^\star) \bigr\|_2^2
\;\leq\; C_4'\, \frac{\| G^{(c)}/N - G^{(t)} \|_{\mathrm{op}}^2}{\sigma_{\min}^2}
\;+\; C_4''\, \frac{\| h^{(c)}/N - h^{(t)} \|_2^2}{\sigma_{\min}^2}.
\end{equation}

Combining (i)--(iii), passing the inner expectation through $Q_S, K_S$
using boundedness from (B1), and absorbing constants into $C_3, C_4$ gives
\eqref{eq:attn-shift}. \hfill$\square$

\paragraph{Interpretation.}
The decomposition mirrors Proposition~\ref{prop:mlp-shift}. The first term $2\,J_{\mathcal{D}_t}^\star$
is the irreducible bilinear residual on $\mathcal{D}_t$ (Proposition~\ref{prop:attn-exact}
applied to deployment) and is the floor any layer-local method must pay.
The second term decays as $d_h'^2 / N$ and gives sample-complexity
guidance specific to attention: the per-head compensator has only
$d_h'^2$ free parameters (e.g., $32^2 = 1024$ or $64^2 = 4096$), so the
calibration budget required for stable estimation is much smaller than for
MLP compensation, where $|S|$ can reach the thousands. This explains
empirically why attention-only pruning is more stable than MLP-only pruning
at small calibration sizes (Table~\ref{tab:calib_sweep}). The third term grows linearly in the
operator-norm gap of the empirical Gram tensors and provides a controlled
sensitivity to calibration--deployment mismatch.

\section{Detailed Algorithms}
\label{app:algorithms}

This appendix provides detailed pseudocode for the components of CORP referenced in the main text. Algorithm~\ref{alg:CORP} summarizes the full pipeline, and the following algorithms detail each step.

\subsection{MLP Channel Ranking}
\label{app:rank_mlp}

\begin{algorithm}[h]
\caption{RankMLP: Hidden-Channel Ranking for MLP Pruning}
\label{alg:rank_mlp}
\begin{algorithmic}
\STATE {\bfseries Input:} Hidden activations $x\in\mathbb{R}^d$, sparsity $s_{\mathrm{mlp}}$
\STATE {\bfseries Output:} Kept indices $S$, pruned indices $P$

\FOR{$j=1$ {\bfseries to} $d$}
  \STATE Compute score $r_j \leftarrow \mathbb{E}[x_j^2]$ \COMMENT{activation-based}
\ENDFOR

\STATE Select top $(1-s_{\mathrm{mlp}})d$ indices by $r_j$ as $S$
\STATE $P \leftarrow \{1,\dots,d\} \setminus S$
\STATE {\bfseries return} $(S,P)$
\end{algorithmic}
\end{algorithm}

Algorithm~\ref{alg:rank_mlp} specifies the hidden-channel ranking procedure used
for MLP pruning.
The ranker scores channels using activation statistics computed on the calibration set
and selects channels to preserve based on the target sparsity.

\subsection{MLP Affine Compensation}
\label{app:comp_mlp}

\begin{algorithm}[h]
\caption{CompensateMLP: Affine Compensation for Pruned MLP Channels}
\label{alg:comp_mlp}
\begin{algorithmic}
\STATE {\bfseries Input:} Layer $y=Wx+b$, activations $x$, indices $(S,P)$, ridge $\lambda$
\STATE {\bfseries Output:} Compensated parameters $(\widehat W_S,\widehat b)$

\STATE Form calibration matrices $X_S,X_P$ from $x_S,x_P$
\STATE Compute $\mu_S=E[x_S]$, $\mu_P=E[x_P]$
\STATE Compute $\Sigma_{SS}=E[(x_S-\mu_S)(x_S-\mu_S)^\top]$
\STATE Compute $\Sigma_{PS}=E[(x_P-\mu_P)(x_S-\mu_S)^\top]$

\STATE $B \leftarrow \Sigma_{PS}(\Sigma_{SS}+\lambda I)^{-1}$
\STATE $c \leftarrow \mu_P - B\mu_S$

\STATE $\widehat W_S \leftarrow W_S + W_P B$
\STATE $\widehat b \leftarrow b + W_P c$

\STATE {\bfseries return} $(\widehat W_S,\widehat b)$
\end{algorithmic}
\end{algorithm}

Algorithm~\ref{alg:comp_mlp} details the closed-form affine compensation used to
correct the bias introduced by MLP channel pruning.
The procedure estimates second-order statistics from calibration activations and
folds the compensation directly into the output projection weights.

\subsection{Attention Dimension Ranking}
\label{app:rank_attn}
\begin{algorithm}[h]
\caption{RankAttn: Logit-Energy Ranking for Attention Pruning}
\label{alg:rank_attn}
\begin{algorithmic}
\STATE {\bfseries Input:} Queries $Q$, keys $K$, sparsity $s_{\mathrm{attn}}$
\STATE {\bfseries Output:} Kept indices $S$, pruned indices $P$

\FOR{$j=1$ {\bfseries to} $d_h$}
  \STATE $s_j \leftarrow \mathbb{E}[\|q_j\|_2^2 \|k_j\|_2^2]$
\ENDFOR

\STATE Select top $(1-s_{\mathrm{attn}})d_h$ indices by $s_j$ as $S$
\STATE $P \leftarrow \{1,\dots,d_h\} \setminus S$
\STATE {\bfseries return} $(S,P)$
\end{algorithmic}
\end{algorithm}

Algorithm~\ref{alg:rank_attn} presents the logit-energy based ranking for attention
projection pruning.
Each head dimension is scored by its expected contribution to the attention logits,
allowing structured pruning without modifying the attention layout.

\subsection{Attention Logit Compensation}
\label{app:comp_attn}
\begin{algorithm}[h]
\caption{CompensateAttn: Logit-Space Compensation for Attention Pruning}
\label{alg:comp_attn}
\begin{algorithmic}
\STATE {\bfseries Input:} Projections $(W_Q,W_K)$, queries $Q$, keys $K$, indices $(S,P)$, ridge $\lambda$
\STATE {\bfseries Output:} Compensated projections $(\hat W_{Q,S},\hat W_{K,S})$

\STATE Split $Q=[Q_S\ Q_P]$, $K=[K_S\ K_P]$

\STATE Initialize $G \leftarrow 0$, $h \leftarrow 0$
\FOR{each calibration sample $b$}
    \STATE $G \leftarrow G + (K_{S,b}^\top K_{S,b}) \otimes (Q_{S,b}^\top Q_{S,b})$
    \STATE $h \leftarrow h + \mathrm{vec}\!\left((Q_{S,b}^\top Q_{P,b})(K_{P,b}^\top K_{S,b})\right)$
\ENDFOR

\STATE Solve $(G + \lambda I)\,\mathrm{vec}(M) = h$
\STATE $M \leftarrow \mathrm{mat}(\mathrm{vec}(M))$

\STATE Compute SVD $I+M=U\Sigma V^\top$

\STATE $\hat W_{Q,S} \leftarrow W_{Q,S} U\Sigma^{1/2}$
\STATE $\hat W_{K,S} \leftarrow W_{K,S} V\Sigma^{1/2}$

\STATE {\bfseries return} $(\hat W_{Q,S},\hat W_{K,S})$
\end{algorithmic}
\end{algorithm}

Algorithm~\ref{alg:comp_attn} describes the logit-space compensation scheme for
attention pruning.
The method recovers missing logit contributions by solving a small Sylvester equation
and folding the solution into the retained query and key projections.

\begin{figure}[t]
    \centering
    \includegraphics[width=\columnwidth]{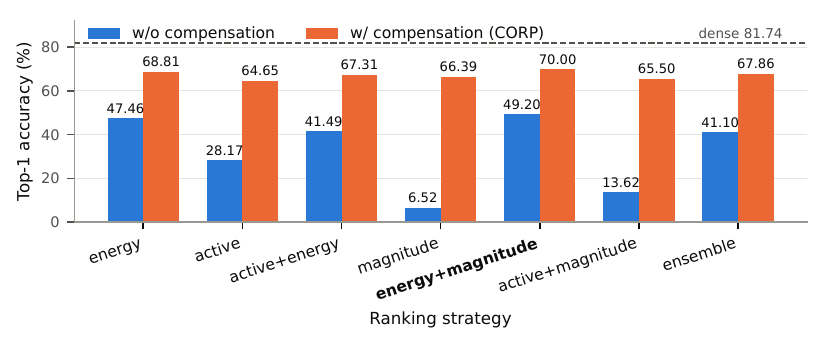}
    \caption{Comparison of ranking strategies with and without compensation. Activation-based and magnitude-based criteria show varied performance, while combined strategies achieve higher accuracy. Compensation consistently improves results across all ranking methods, highlighting that representation recovery is more critical than the choice of ranking.}
    \label{fig:ranking_ablation}
\end{figure}
\section{Ranking ablation}
We compare activation energy, magnitude-based, active probability and combined ranking strategies from Sec.~\ref{sec:ranking}, with and w/o compensation on DeiT-Base at 50\% joint sparsity. 

The active ranking policy scores each channel by its activation frequency, $P(|x| > \epsilon)$, estimated over the calibration set. Channels that activate less frequently are assigned higher pruning priority. The intuition is that rarely activated channels contribute little to downstream representations, so removing them introduces minimal error regardless of weight magnitude.
The combination of activation energy, magnitude-based strategy achieves the best accuracy and is used by default.
The result is in Figure \ref{fig:ranking_ablation}
 
\label{app:ranking}

\section{Additional Analyses}
\label{app:additional}

\subsection{Empirical Recoverability}
\label{app:rho2}

Propositions~\ref{prop:mlp-gain} and~\ref{prop:attn-gain} define $\rho^2_{W_P}$ and $\rho^2_{\mathrm{attn}}$ as the fraction of the pruned contribution's energy that is linearly recoverable from the retained components.
We measure both on DeiT-Large at 50\% joint MLP and attention sparsity with the default energy--magnitude ranking and $\lambda=10^{-3}$.
The calibration and evaluation sets are disjoint random samples of 4000 ImageNet training images each, both drawn from all 1000 classes.
For a unit $u$ (an MLP layer or an attention head), let $E_u$ denote the energy of the removed contribution and $R_u$ the residual energy after compensation, so that $\rho_u^2 = 1 - R_u/E_u$ is the fraction of that unit's removed energy recovered by linear compensation.
We report the \emph{pooled} value
\begin{equation}
\rho^2_{\mathrm{pool}} \;=\; 1 - \frac{\sum_u R_u}{\sum_u E_u} \;=\; \frac{\sum_u E_u\,\rho_u^2}{\sum_u E_u},
\end{equation}
which is the fraction of the \emph{total} pruned energy recovered and therefore the quantity that corresponds to the representation distortion removed by CORP, together with the unweighted mean, standard deviation, and range over units, which show how uniformly compensation works.
Three protocols are used:
\begin{itemize}
\item \textbf{Calibration}: fit and evaluate the compensator on the calibration set (in-sample recoverability).
\item \textbf{Evaluation}: fit and evaluate on the evaluation set (an oracle upper bound that is not available to CORP).
\item \textbf{Transfer}: fit the compensator once on the calibration set, freeze it, and evaluate on the held-out evaluation set. This matches CORP's deployment; the gap between evaluation and transfer is the cost of fitting only once.
\end{itemize}

\begin{figure}[h]
\centering
\includegraphics[width=\textwidth]{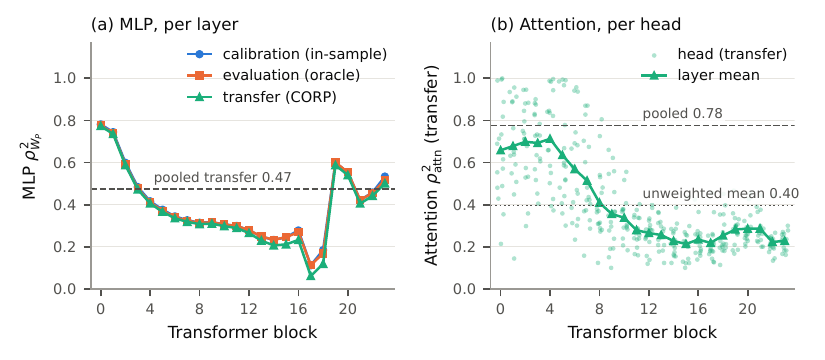}
\caption{Layer-wise and head-wise recoverability on DeiT-Large (50\% joint sparsity).
\textbf{(a)} MLP $\rho^2_{W_P}$ per block under the three protocols; the three curves nearly coincide, so fitting once on the calibration set loses little.
\textbf{(b)} Transfer $\rho^2_{\mathrm{attn}}$ of each of the 384 heads (dots) and the per-block mean (line); the gap between the energy-weighted pooled value and the unweighted mean reflects the concentration of removed logit energy in a few highly recoverable heads.}
\label{fig:rho2_layerwise}
\end{figure}

Table~\ref{tab:rho2_summary} summarizes the results and Figure~\ref{fig:rho2_layerwise} shows the per-layer and per-head distributions.
Fitting the compensator once costs almost nothing relative to the oracle: the pooled oracle-to-transfer gap is 0.0007 for attention ($0.7775\to0.7768$) and 0.0089 for the MLP ($0.4825\to0.4736$).
The pooled and unweighted statistics differ because the removed attention-logit energy is highly concentrated: the top 10 of 384 heads account for 56.4\% of the removed energy and have a mean $\rho^2$ of 0.980, so the pooled value is dominated by the most influential heads.
Quoting only the pooled value would overstate how uniformly compensation works, while quoting only the unweighted mean would understate how much representation distortion it removes.
Recoverability also decreases with depth for both module types (Figure~\ref{fig:rho2_layerwise}), and the largest transfer gaps occur in the deeper MLP blocks.

\begin{table}[h]
\centering
\small
\setlength{\tabcolsep}{5pt}
\begin{tabular}{llcccc}
\toprule
Protocol & Module & Pooled $\rho^2$ & Mean $\pm$ std. & Min & Max \\
\midrule
Calibration & MLP (24 layers)       & 0.4893 & 0.3944 $\pm$ 0.1707 & 0.1126 & 0.7816 \\
Evaluation (oracle) & MLP           & 0.4825 & 0.3909 $\pm$ 0.1698 & 0.1133 & 0.7766 \\
\textbf{Transfer (CORP)} & MLP      & \textbf{0.4736} & 0.3755 $\pm$ 0.1789 & 0.0619 & 0.7735 \\
\midrule
Calibration & Attention (384 heads) & 0.7825 & 0.4018 $\pm$ 0.2336 & 0.1045 & 0.9995 \\
Evaluation (oracle) & Attention     & 0.7775 & 0.3993 $\pm$ 0.2323 & 0.1018 & 0.9995 \\
\textbf{Transfer (CORP)} & Attention & \textbf{0.7768} & 0.3970 $\pm$ 0.2335 & 0.1008 & 0.9995 \\
\bottomrule
\end{tabular}
\caption{Recoverability $\rho^2$ on DeiT-Large (50\% joint sparsity) under the three protocols.
Pooled values are energy weighted; mean, standard deviation, minimum, and maximum are unweighted over units.}
\label{tab:rho2_summary}
\end{table}

\paragraph{Content dependence.}
Because $M$ is fitted once from aggregate statistics while attention patterns are content dependent, we repeat the transfer protocol under a stricter split: the compensator is fitted on images from 500 ImageNet classes and evaluated on the remaining 500 classes, with 2000 images per side, so the fitted and evaluated inputs share no object category.
A same-distribution control with the same sample size isolates the effect of the smaller sample.
Table~\ref{tab:rho2_classdisjoint} shows that the attention gap under the class-disjoint split (0.0016) is within 0.0004 of the control (0.0012), so almost all of the degradation comes from finite-sample estimation rather than content shift; MLP compensation is more sensitive (0.0233 vs.\ 0.0179).
These results support transfer to unseen inputs and unseen object categories from the same deployment distribution; they do not establish robustness under arbitrary distribution shift.

\begin{table}[h]
\centering
\small
\setlength{\tabcolsep}{5pt}
\begin{tabular}{llcccc}
\toprule
Split & Module & Oracle (eval) & Transfer & Gap & Transfer mean $\pm$ std. \\
\midrule
Class-disjoint (500/500 classes) & MLP       & 0.4771 & 0.4538 & 0.0233 & 0.3437 $\pm$ 0.1810 \\
Class-disjoint (500/500 classes) & Attention & 0.7795 & 0.7779 & 0.0016 & 0.3942 $\pm$ 0.2352 \\
Same-distribution control        & MLP       & 0.4866 & 0.4687 & 0.0179 & 0.3674 $\pm$ 0.1838 \\
Same-distribution control        & Attention & 0.7779 & 0.7767 & 0.0012 & 0.3973 $\pm$ 0.2339 \\
\bottomrule
\end{tabular}
\caption{Pooled $\rho^2$ under a class-disjoint calibration/evaluation split and a same-distribution control of matched sample size (2000 images per side). Gap is oracle minus transfer.}
\label{tab:rho2_classdisjoint}
\end{table}

\subsection{Ridge Coefficient Sensitivity}
\label{app:lambda}

We use one fixed value, $\lambda=10^{-3}$, for every MLP and attention solve in all reported experiments; $\lambda$ is not tuned per model, layer, head, or sparsity ratio, and the ridge term only regularizes each covariance or normal-equation solve.
Table~\ref{tab:lambda_sweep} reports Top-1 accuracy on DeiT-Large under 50\%, 60\%, and 70\% joint sparsity for $\lambda\in\{10^{-5},10^{-4},10^{-3},10^{-2}\}$.
Over $10^{-5}$--$10^{-3}$ the standard deviation stays below 0.06 points at every sparsity level, including 70\%, where the retained covariance is expected to be less well conditioned.
The larger value $10^{-2}$ reduces accuracy slightly (by 0.66 points at 70\%), which indicates mild over-regularization rather than numerical instability.
CORP therefore does not require setting-specific tuning of $\lambda$; we selected $10^{-3}$ as a single conservative default.

\begin{table}[h]
\centering
\small
\setlength{\tabcolsep}{6pt}
\begin{tabular}{lccccc}
\toprule
Joint sparsity & $\lambda=10^{-5}$ & $10^{-4}$ & $10^{-3}$ (default) & $10^{-2}$ & Mean $\pm$ std.\ ($10^{-5}$--$10^{-3}$) \\
\midrule
50\% & 80.34 & 80.33 & 80.30 & 80.06 & 80.32 $\pm$ 0.02 \\
60\% & 75.95 & 75.96 & 75.92 & 75.57 & 75.94 $\pm$ 0.02 \\
70\% & 64.49 & 64.49 & 64.40 & 63.74 & 64.46 $\pm$ 0.05 \\
\bottomrule
\end{tabular}
\caption{Top-1 accuracy of DeiT-Large under joint sparsity as a function of the ridge coefficient $\lambda$.}
\label{tab:lambda_sweep}
\end{table}

\section{Efficiency} 
\label{app:eff}

\begin{table*}[h]
\centering
\resizebox{\textwidth}{!}{
\begin{tabular}{c c c c c c c c c c}
\hline
Model & Sparsity & Top-1 & Param & FLOPs & Lat & TP & Param$\downarrow$ & FLOPs$\downarrow$ & TP$\uparrow$ \\
 &  & (\%) & (M) & (G) & (ms) & (fps) & (\%) & (\%) & (×) \\
\hline

\multirow{9}{*}{DeiT-T}
& 0.0 & 72.02 & 5.7 & 1.4 & 3.79 & 3514 & 0.0 & 0.0 & 1.00 \\
& 0.1 & 71.26 & 5.3 & 1.2 & 4.95 & 3247 & 7.6 & 19.0 & 0.92 \\
& 0.2 & 69.15 & 4.8 & 1.1 & 5.01 & 3453 & 15.3 & 25.6 & 0.98 \\
& 0.3 & 64.46 & 4.4 & 1.0 & 4.84 & 3554 & 23.2 & 32.4 & 1.01 \\
& 0.4 & 55.73 & 4.0 & 0.9 & 4.89 & 3699 & 30.9 & 39.0 & 1.05 \\
& 0.5 & 41.36 & 3.5 & 0.8 & 4.91 & 4048 & 38.8 & 45.9 & 1.15 \\
& 0.6 & 21.02 & 3.1 & 0.7 & 5.06 & 4136 & 46.4 & 52.4 & 1.18 \\
& 0.7 & 2.33  & 2.6 & 0.6 & 5.03 & 4434 & 54.1 & 59.0 & 1.26 \\

\hline
\multirow{9}{*}{DeiT-S}
& 0.0 & 79.72 & 22.1 & 5.0 & 3.71 & 1370 & 0.0 & 0.0 & 1.00 \\
& 0.1 & 78.48 & 20.3 & 4.2 & 4.96 & 1334 & 7.9 & 14.5 & 0.97 \\
& 0.2 & 76.78 & 18.6 & 3.9 & 5.25 & 1440 & 15.9 & 21.7 & 1.05 \\
& 0.3 & 74.53 & 16.8 & 3.5 & 4.99 & 1499 & 24.0 & 29.3 & 1.09 \\
& 0.4 & 70.84 & 15.0 & 3.2 & 4.96 & 1621 & 32.0 & 36.5 & 1.18 \\
& 0.5 & 58.37 & 13.2 & 2.8 & 4.91 & 1831 & 40.2 & 44.1 & 1.34 \\
& 0.6 & 34.54 & 11.4 & 2.4 & 5.04 & 1917 & 48.1 & 51.4 & 1.40 \\
& 0.7 & 15.64 & 9.7  & 2.1 & 5.17 & 2108 & 56.1 & 58.6 & 1.54 \\

\hline
\multirow{9}{*}{DeiT-B}
& 0.0 & 81.74 & 86.6 & 18.3 & 4.28 & 418 & 0.0 & 0.0 & 1.00 \\
& 0.1 & 80.26 & 79.6 & 16.2 & 4.94 & 431 & 8.1 & 11.6 & 1.03 \\
& 0.2 & 79.14 & 72.6 & 14.8 & 4.90 & 463 & 16.2 & 19.3 & 1.11 \\
& 0.3 & 76.98 & 65.4 & 13.3 & 5.29 & 502 & 24.5 & 27.3 & 1.20 \\
& 0.4 & 74.23 & 58.4 & 11.9 & 4.91 & 549 & 32.6 & 35.0 & 1.31 \\
& 0.5 & 70.00 & 51.2 & 10.4 & 4.91 & 630 & 40.9 & 43.0 & 1.51 \\
& 0.6 & 63.33 & 44.2 & 9.0  & 4.88 & 685 & 49.0 & 50.7 & 1.64 \\
& 0.7 & 49.78 & 37.2 & 7.6  & 4.90 & 766 & 57.1 & 58.4 & 1.83 \\

\hline
\multirow{9}{*}{DeiT-L}
& 0.0 & 84.58 & 304.4 & 63.5 & 12.96 & 121 & 0.0 & 0.0 & 1.00 \\
& 0.1 & 84.57 & 279.5 & 56.6 & 12.43 & 129 & 8.2 & 10.8 & 1.06 \\
& 0.2 & 84.35 & 254.7 & 51.6 & 11.41 & 140 & 16.3 & 18.7 & 1.15 \\
& 0.3 & 83.61 & 229.0 & 46.5 & 10.62 & 151 & 24.8 & 26.8 & 1.25 \\
& 0.4 & 82.75 & 204.1 & 41.5 & 10.66 & 166 & 33.0 & 34.7 & 1.37 \\
& 0.5 & 80.30 & 178.5 & 36.3 & 10.61 & 187 & 41.4 & 42.8 & 1.54 \\
& 0.6 & 75.92 & 153.6 & 31.4 & 10.57 & 207 & 49.5 & 50.6 & 1.71 \\
& 0.7 & 64.40 & 128.8 & 26.4 & 10.58 & 238 & 57.7 & 58.5 & 1.96 \\

\hline
\multirow{9}{*}{DeiT-H}
& 0.0 & 84.97 & 632.1 & 172.8 & 30.84 & 43 & 0.0 & 0.0 & 1.00 \\
& 0.1 & 84.96 & 579.7 & 153.7 & 27.70 & 48 & 8.3 & 11.1 & 1.12 \\
& 0.2 & 84.91 & 527.2 & 139.9 & 24.82 & 52 & 16.6 & 19.0 & 1.21 \\
& 0.3 & 84.66 & 474.8 & 126.2 & 22.73 & 57 & 24.9 & 27.0 & 1.32 \\
& 0.4 & 84.23 & 422.3 & 112.4 & 20.70 & 63 & 33.2 & 35.0 & 1.46 \\
& 0.5 & 83.27 & 369.9 & 98.7  & 19.16 & 71 & 41.5 & 42.9 & 1.64 \\
& 0.6 & 80.98 & 317.4 & 84.9  & 16.83 & 78 & 49.8 & 50.9 & 1.82 \\
& 0.7 & 75.38 & 265.0 & 71.2  & 14.71 & 92 & 58.1 & 58.8 & 2.13 \\

\hline
\end{tabular}}
\caption{Accuracy and efficiency trade-offs of CORP across model scales. 
We report Top-1 accuracy, parameters (M), FLOPs (G), latency (ms), and throughput (fps).
FLOPs and throughput are measured on a single RTX 3090 GPU with batch sizes 256/128/64/32/16 for Tiny/Small/Base/Large/Huge. Latency uses batch size 1.}
\label{tab:efficiency_all}
\end{table*}

\end{document}